\documentclass[10pt,twocolumn,letterpaper]{article}
\usepackage[pagenumbers]{cvpr}
\usepackage{graphicx,amsmath,amssymb,amsthm,booktabs}
\definecolor{cvprblue}{rgb}{0.21,0.49,0.74}
\usepackage[pagebackref,breaklinks,colorlinks,allcolors=cvprblue]{hyperref}
\hypersetup{pdftitle={RALC: Risk-Controlled Local Adoption of Segmentation Updates with Few Calibration Images},pdfauthor={Shize Yin}}
\newtheorem{theorem}{Theorem}
\newcommand{\E}{\mathbb E}
\newcommand{\pos}[1]{[#1]_+}
\title{RALC: Risk-Controlled Local Adoption of Segmentation Updates with Few Calibration Images}
\author{Shize Yin\\Northeastern University, China}
\begin{document}
\maketitle
\raggedbottom
\begin{abstract}Fine-tuning a road-scene segmenter for rain and nighttime driving can introduce errors on normal daytime images. With few labeled normal images for calibration, a whole-image risk check may reject changes that would pass if adopted locally. Risk-Controlled Adoption of Label Changes (RALC) selects local changes from a fixed model update, using confidence differences and an image score to form nested candidates. On each normal calibration image, RALC measures how many more errors each candidate makes than the original model, assigning zero loss when errors stay the same or fall. At test time, each replacement can add at most one error, so the replacement count bounds the net increase without test labels. RALC combines this candidate-dependent bound with recorded calibration losses to select the largest set that passes the risk check. If the models and adoption rules are fixed before calibration and normal calibration and test images are exchangeable, the expected per-image loss stays within budget. On held-out BDD100K, we evaluated three weight-interpolated updates with 10 additional normal images per calibration draw. At a common fixed budget, adverse-scene mIoU was 1.89--3.19 percentage points higher with RALC than with whole-image adoption, while measured mean normal loss was also higher. After matching this loss post hoc at 1,000 pixels per image, RALC still led by 1.135--2.428 points. Selecting changes in $8{\times}8$-pixel cells recovered 93--98\% of this advantage. With 250 calibration images at the original budget, both rules adopted all three updates in full.
\end{abstract}
\section{Introduction}
Fine-tuning a road-scene segmenter for rain, snow or darkness can also change what it predicts in ordinary daytime traffic. Within a single image, the update may correct one pixel and overwrite a label that was already right. Whole-image adoption accepts or rejects all changes in an image together. If only a few labeled normal images are available for calibration, which changes can we keep while limiting the increase in errors on normal scenes?

With both models fixed, we limit expected normal-scene loss, with each image contributing its net increase in errors or zero if errors fall. A finite-sample check combines calibration losses with an upper bound on unknown test loss. Whole-image adoption must allow for every disagreement, so with few calibration images it can reject changes that would pass locally.

\begin{figure*}[t]
\centering\includegraphics[width=\textwidth]{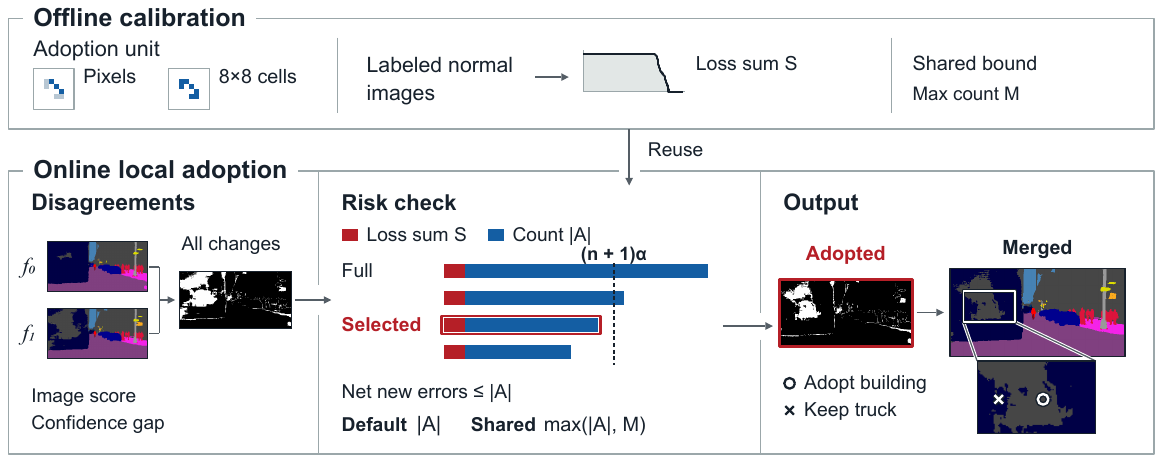}
\caption{Offline calibration stores sums $S$ of the upper-envelope losses defined in Sec.~3.3. Online adoption adds the current candidate's replacement count $|A|$. Pixels are the default unit; grid candidates group disagreements within each cell. Bars show four of the 129 shared levels, from largest to smallest candidate on the full test image. The first whose sum is at most $(n+1)\alpha$ supplies the outlined mask. The shared-bound variant also stores the largest calibration replacement count $M$ per level. Both marked pixels belong to the truck. RALC adopts the proposed \emph{building} label at the circle and retains the \emph{truck} label at the cross.}
\label{fig:method}
\end{figure*}

RALC (Risk-Controlled Adoption of Label Changes) bounds the unknown loss by the number of proposed label replacements (Fig.~\ref{fig:method}). Reducing this set tightens the bound and can admit local changes when whole-image adoption fails. For a configuration fixed before calibration and exchangeable normal calibration and test images, conformal risk control (CRC)~\cite{crc} bounds expected per-image loss over both the calibration draw and the future normal image.

We fixed the primary adoption rules before evaluating them on BDD100K. Across three weight-interpolated updates, with ten additional normal calibration images per draw, RALC gained 1.89--3.19 percentage points (pp) in adverse-scene mIoU over whole-image adoption at a common budget, while incurring higher measured mean normal loss. Matching that mean post hoc at 1,000 pixels per image left an advantage of 1.135--2.428 pp (Table~\ref{tab:small}). The gain also persisted against learned and randomized whole-image selectors at matched mean normal loss. Selecting changes in $8{\times}8$-pixel cells recovered 93--98\% of the advantage over Whole at matched mean normal loss (Fig.~\ref{fig:riskbenefit}, bottom).

Our contributions are as follows:
\begin{itemize}
\item We formulate local adoption of a fixed segmentation update under limited normal calibration data and analyze how the available evidence restricts the extent of adoption.
\item We propose RALC, which combines the candidate replacement-count bound with calibration losses to select the largest feasible set in a nested family, with a finite-sample guarantee on expected normal-image regression loss under CRC.
\item At matched measured mean normal loss, we show that within-image adoption retains gains left unused by whole-image selection. Small grids and pixels both realize this benefit, clarifying the roles of adoption granularity and calibration size.
\end{itemize}

\section{Related work}
\paragraph{Model updates and backward compatibility.}
Positive-Congruent Training gives greater distillation weight to examples the original model predicts correctly, reducing changes from correct to incorrect predictions after an update~\cite{pct}. Fine-tuning can distort pretrained features~\cite{distort}, while weight interpolation and averaging can improve robustness~\cite{wise,soups}. Test-time adaptation adjusts model parameters on unlabeled inputs through entropy minimization, parameter restoration or sample selection~\cite{tent,cotta,eata,sar}. For fixed classification models, Tr{\"a}uble et al.~\cite{trauble2021updates} use probabilistic estimates to decide which examples to re-evaluate and whether to adopt changed predictions. RALC considers the adoption decision within a segmentation map, using labeled normal images to calibrate it.

\paragraph{Selective prediction and risk control.}
Selective prediction allows abstention on some inputs, trading the fraction of inputs receiving predictions against their error rate~\cite{selectivenet}. CRC, which supplies RALC's expected-loss criterion, calibrates a family of predictions to control expected task loss under exchangeability~\cite{crc}. Risk-controlling prediction sets use upper confidence bounds on risk, and Learn then Test uses hypothesis testing to identify acceptable parameter settings~\cite{rcps,ltt}. Automatically Adaptive CRC adapts risk control to sample difficulty to obtain approximate conditional guarantees~\cite{aacrc}. ReCIRC estimates a risk curve for each input, finds thresholds corresponding to a common predicted risk, and applies CRC to the resulting family~\cite{recirc}.

\paragraph{Risk calibration in segmentation.}
Wang et al.\ use prediction-disagreement geometry to control test-time adaptation; their regional gate calibrates a fixed-denominator harmful-area surrogate~\cite{wang}. RALC calibrates the positive net increase in pixel errors from a fixed update, using each candidate's replacement count as the test-loss bound. Shrinking the candidate therefore reduces the reserve needed by the finite-sample check. Luo et al.\ learn image-adaptive thresholds to improve image-conditional coverage of segmentation prediction sets~\cite{luo}.

\section{Risk-controlled local adoption}
\subsection{Task and control objective}
The original model $f_0$ and updated model $f_1$ are fixed before calibration. On image $x$, their predicted labels at pixel $p$ are $\hat y^0_p$ and $\hat y^1_p$. They disagree on $D(x)=\{p:\hat y^0_p\ne\hat y^1_p\}$. Let $\lambda$ index candidate sets of changes. For $A(x,\lambda)\subseteq D(x)$, the merged prediction uses updated labels on $A(x,\lambda)$ and original labels elsewhere. Given ground truth $y$, let $e(x,y,A)$ count its errors on valid annotated pixels. The original prediction corresponds to $A=\varnothing$, giving the per-image regression loss
\begin{equation}
 H(x,y,\lambda)=\pos{e(x,y,A(x,\lambda))-e(x,y,\varnothing)} ,
 \label{eq:loss}
\end{equation}
where $[a]_+=\max(a,0)$. Each normal image contributes the errors in excess of corrections, with zero loss when errors stay the same or fall. We seek $\E[H]\leq\alpha$, where the budget $\alpha$ is in pixels per image.

\subsection{Local candidates and their count bound}
Each label replacement can add at most one error. Unchanged pixels have identical errors under the original and merged predictions. Thus $B(x,\lambda)=|A(x,\lambda)|$ bounds $H(x,y,\lambda)$ without test labels. We count all proposed replacements, including positions without valid annotations, which preserves the bound.

To construct candidates, we rank disagreeing pixels by how much more confident the updated model is in its proposed label than the original model was in its own. For pixel $p$, the gap is
\begin{equation}
 r_p=p_1(\hat y^1_p\mid x)-p_0(\hat y^0_p\mid x),\qquad p\in D(x).
\end{equation}
We place the gaps in 32 equal-width bins. At each shared level $\lambda$, the fixed image score $s(x)\in[0,1]$ determines that image's bin cutoff; higher scores permit lower cutoffs. Disagreements at or above the cutoff form the candidate. As $\lambda$ increases, cutoffs do not decrease, so $A(x,\lambda')\subseteq A(x,\lambda)$ for $\lambda'\geq\lambda$. The 129 shared levels accommodate these image-specific cutoffs, from all disagreements to none; several levels can select the same pixels on one image (Sec.~S15). We train the scorer by logistic regression before calibration to predict whether full adoption reduces an image's error count (Sec.~S6). Omitting it gives every image the same cutoff at a given level.

For grid candidates, each cell takes the mean gap of its disagreeing pixels. Binning this mean makes those pixels enter or leave together. The risk check still counts individual replacements.

\subsection{Offline calibration and online adoption}
\paragraph{Offline calibration.}
We evaluate every candidate on each labeled normal image. The loss $H$ need not decrease as the proposal shrinks. A candidate that corrects one error and introduces another has zero loss; removing the correction leaves a loss of one. We therefore use CRC's monotone upper envelope~\cite{crc}. For $n$ calibration images, we store
\begin{align}
 L_i(\lambda)&=\max_{\lambda'\geq\lambda} H(x_i,y_i,\lambda'),&
 S_n(\lambda)&=\sum_{i=1}^n L_i(\lambda).
\end{align}
Nesting ensures that each more conservative candidate changes a subset of the current pixels, so $H\leq L\leq B$.

\paragraph{Online adoption.}
On a test image, we construct the same candidates and count $B(x,\lambda)$ at each level. Reusing the stored calibration sum $S_n(\lambda)$, we select
\begin{equation}
 \hat\lambda(x)=\min\{\lambda:S_n(\lambda)+B(x,\lambda)\leq(n+1)\alpha\}.
 \label{eq:rule}
\end{equation}
Scanning from the largest candidate to the smallest, the first passing level gives the largest admissible set (Fig.~\ref{fig:method}). We use updated labels on that set and original labels elsewhere. Complete rollback sets both terms to zero and is always feasible.

\paragraph{Shared-bound variant.}
The shared variant keeps the pixel ordering, candidate sets and calibration loss sums of the default rule. During calibration it also stores the maximum replacement count at each level, $M_n(\lambda)=\max_{1\le i\le n}B(x_i,\lambda)$. In Eq.~\eqref{eq:rule} it replaces $B(x,\lambda)$ by $\max\{B(x,\lambda),M_n(\lambda)\}$. A calibration image with many proposed changes can therefore restrict adoption on another image with few. At a fixed budget, every candidate that passes this shared check also passes the default check.

\subsection{Risk guarantee and sample size}
Let $T$ denote the models, scoring rule and candidate construction fixed before calibration, and let $Z_i=(x_i,y_i)$ denote a labeled normal image. The risk guarantee applies to a finite nested family containing complete rollback, including the grid construction above.
\begin{theorem}\label{thm:validity}
Fix the configuration $T$ and budget $\alpha\geq0$. Let the normal calibration samples and the future normal test sample be $Z_1,\ldots,Z_{n+1}$, whose joint distribution is exchangeable given $T$. Suppose the finite nested adoption family contains complete rollback. Under these conditions, Eq.~\eqref{eq:rule} satisfies
\[
\E[H_{n+1}(\hat\lambda)\mid T]\leq\E[L_{n+1}(\hat\lambda)\mid T]\leq\alpha .
\]
\end{theorem}
\begin{proof}
Let $\lambda^\star$ be the smallest level for which the average of the $n+1$ actual envelope losses is at most $\alpha$. Complete rollback ensures this level exists. At the deployed level, replacing the test count bound by the actual test envelope can only reduce the sum. The deployed level therefore passes the reference check, giving $\lambda^\star\leq\hat\lambda$. Monotonicity gives $L_{n+1}(\hat\lambda)\leq L_{n+1}(\lambda^\star)$. The reference level is symmetric in all $n+1$ samples. By exchangeability conditional on $T$, the expected test envelope at that level equals the expected average envelope, which is at most $\alpha$. Finally $H\leq L$ pointwise.
\end{proof}

The expectation averages over the calibration draw and the future normal image. Individual images can exceed the budget. Since $S_n\geq0$, a feasible candidate changes at most $\min(|D(x)|,\lfloor(n+1)\alpha\rfloor)$ pixels. Whole-image adoption therefore requires $|D(x)|\leq(n+1)\alpha$ even if every calibration loss is zero. The same check can still admit a local subset.

The same guarantee holds for the shared variant under the theorem's conditions. Its bound dominates $B(x,\lambda)$ at every level, so it selects a subset of the default set and its monotone envelope loss is no larger.

\section{Experiments}
\begin{table*}[t]
\caption{Adverse mIoU gain (pp) on 500 BDD images, with ten normal calibration images per draw. Full update adopts every change; $\overline H$ is mean normal loss on 250 held-out images (pixels/image). At fixed budget, $\rho=\overline H/\alpha$ and $\Delta_{\rm R-W}$ is RALC minus Whole. Panel (b) matches loss post hoc for WI updates. Rand.\ is randomized whole-image adoption; ReCIRC is the instance variant. WI denotes weight interpolation and FT fine-tuning.}
\label{tab:small}
\centering
\begingroup\small\setlength{\tabcolsep}{4pt}\renewcommand{\arraystretch}{1.13}
\centering
\begin{tabular*}{\textwidth}{@{\extracolsep{\fill}}llrrrrrrrrr@{}}
\toprule
\multicolumn{2}{l}{\textbf{(a) Fixed budget}} & \multicolumn{2}{c}{Full update} & \multicolumn{7}{c}{Calibrated adoption ($\tau=0.005$)} \\
\cmidrule(lr){3-4}\cmidrule(l){5-11}
Model & Update & Gain & $\overline H$ & \multicolumn{2}{c}{Whole} & \multicolumn{2}{c}{Shared} & \multicolumn{2}{c}{RALC} & RALC $-$ Whole \\
 & & (pp) & (px/img) & Gain & $\rho$ & Gain & $\rho$ & Gain & $\rho$ & (pp) \\
\midrule
B2 & WI & +8.896 & 3,120 & +1.235 & 0.103 & +2.771 & 0.020 & \textbf{+4.427} & 0.149 & +3.192 \\
B0 & WI & +5.852 & 6,359 & +0.492 & 0.110 & +2.281 & 0.108 & \textbf{+3.166} & 0.282 & +2.675 \\
Mask2Former & WI & +2.671 & 2,972 & +0.225 & 0.068 & +1.015 & 0.037 & \textbf{+2.114} & 0.161 & +1.889 \\
\addlinespace[2pt]
B2 & FT & +8.770 & 3,773 & +1.017 & 0.107 & +2.813 & 0.021 & \textbf{+4.261} & 0.168 & +3.244 \\
B0 & FT & +5.113 & 10,769 & +0.165 & 0.097 & +2.191 & 0.117 & \textbf{+2.783} & 0.310 & +2.619 \\
Mask2Former & FT & +1.870 & 4,790 & +0.032 & 0.085 & +0.908 & 0.029 & \textbf{+1.825} & 0.213 & +1.793 \\
\bottomrule
\end{tabular*}

\vspace{6pt}
\begin{tabular*}{\textwidth}{@{\extracolsep{\fill}}lrrrr@{}}
\multicolumn{5}{l}{\textbf{(b) Post hoc loss match} ($\overline H=1{,}000$ pixels/image), WI updates only} \\
\addlinespace[2pt]
 & \multicolumn{4}{c}{Difference in adverse mIoU gain (pp)} \\
\cmidrule(l){2-5}
Model & RALC $-$ Whole & RALC $-$ ReCIRC & $8{\times}8$ grid $-$ Rand. & RALC $-$ Rand. \\
\midrule
B2 & \textbf{+2.428} & +2.431 & +1.945 & +2.001 \\
B0 & \textbf{+1.414} & +1.422 & +1.156 & +1.249 \\
Mask2Former & \textbf{+1.135} & +0.936 & +0.880 & +0.956 \\
\bottomrule
\end{tabular*}
\endgroup

\end{table*}

\subsection{Experimental setup}
\paragraph{Data.}
We developed the rules on Cityscapes and ACDC~\cite{cityscapes,acdc}, with MUSES used for development evaluation~\cite{muses}. We fixed the models, budgets and primary Whole and RALC rules before the first BDD100K evaluation~\cite{bdd}. Follow-up controls used the same frozen predictions. The BDD pool contains one image from each of 1,000 training-split videos, selected using weather metadata and a fixed seed. Of the 500 dry daytime images, half form the calibration pool and half are held out to measure normal loss. The other 500 images measure adverse-scene gain in rain, snow, fog or dry night (supplement, Secs.~S9 and S13).

\paragraph{Updates and calibration budgets.}
We fine-tuned Cityscapes-trained SegFormer-B2 (B2), SegFormer-B0 (B0)~\cite{segformer} and Mask2Former (M2F) with a Swin-B backbone~\cite{mask2former,swin} on 807 adverse ACDC images (FT). A separate pool of 699 images from 599 scene identities supplied the scorer training data. Its adverse sequences were disjoint from those used for fine-tuning.

For weight interpolation (WI), we selected the coefficient of the fine-tuned model on training data by maximizing adverse mIoU subject to the normal loss budget at $\tau=0.005$. The coefficients were 0.9, 0.8 and 0.7 for B2, B0 and M2F (supplement, Secs.~S11 and S12). Only after fixing the update and its scorer did we draw the $n$ additional labeled normal images for calibration.

We fix the reference valid-pixel count $P_{\rm ref}$ on an independent normal training set (supplement, Sec.~S13). At $\tau=0.005$, the budget $\alpha=\tau P_{\rm ref}$ allows an expected loss of about 9,694 pixels per normal image (1.05\% of a $720\times1280$ BDD image). We take the first 10, 50, 100 or 250 images of 50 prespecified calibration orders. All methods share these draws, test images and FP32 predictions. At $n=250$ every order uses the whole pool.

\paragraph{Metrics.}
On adverse scenes, we report mIoU gain over the original segmenter ($\Delta_{\rm adv}$, in percentage points). Within each calibration draw we pool confusion counts across images, then average IoU over classes with nonempty union. On normal scenes, we first count each image's net increase in errors, assigning zero if errors fall, and then average over images. The risk ratio $\rho=\overline H/\alpha$ expresses this measured mean loss as a fraction of the budget. We compute means and differences before rounding.

\paragraph{Comparison rules.}
Whole accepts all disagreements or keeps the original prediction, using the same image scorer as RALC. The learned whole-image selector ranks images by predicted benefit and harm. Local rules can select individual disagreements, connected regions or grid cells. ReCIRC-instance constructs local candidates by inverting a predicted loss curve and uses the same per-image count bound as RALC~\cite{recirc}. We fitted its regressor on 300 normal development images before calibration (supplement, Sec.~S8).

\begin{figure*}[t]
\centering\includegraphics[width=\textwidth]{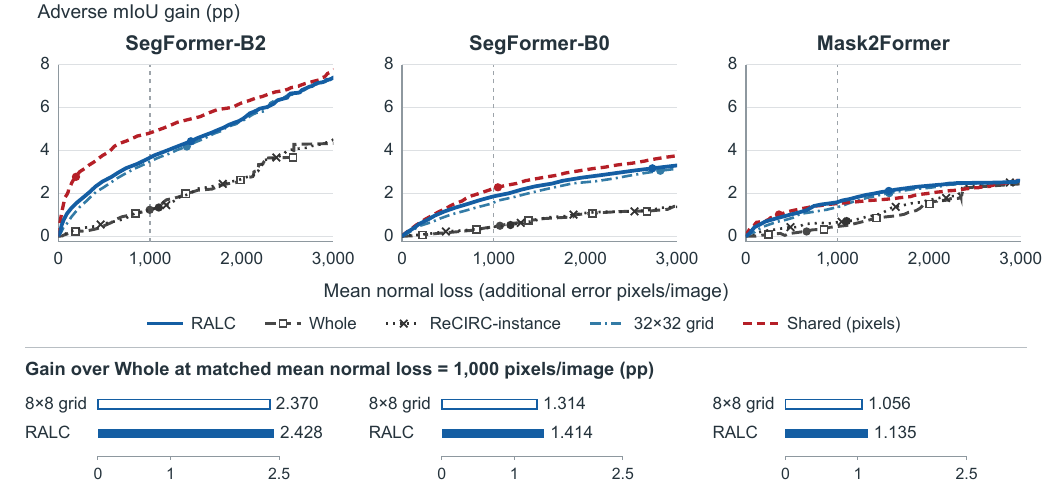}
\caption{Top panels sweep the nominal budget with ten calibration images. Each curve plots adverse mIoU gain against measured normal loss. The grid curve uses $32{\times}32$-pixel cells. Dots mark $\tau=0.005$, squares denote Whole and crosses ReCIRC-instance. Bottom panels compare $8{\times}8$-pixel cells and default RALC with Whole at the dashed line (1,000 pixels/image).}
\label{fig:riskbenefit}
\end{figure*}

\subsection{Performance under limited calibration}
\paragraph{Fixed budget.}
With full adoption, all three WI updates had measured mean loss below the main budget on the held-out normal images (Table~\ref{tab:small}). Whole nevertheless withheld the update on many images when calibrated on just ten. At this budget, RALC exceeded Whole in adverse mIoU by 1.889--3.192 pp for WI and 1.793--3.244 pp for directly fine-tuned models. For B2 WI, adverse mIoU rose from 32.55\% to 33.78\% with Whole, 36.97\% with RALC and 41.44\% with full adoption. Normal mIoU was 46.77\% originally, 48.42\% with Whole and 50.61\% with RALC.

The full B0 FT update incurred 10,769 pixels/image of mean normal loss, above the budget of about 9,694. RALC retained 2.783 pp of adverse gain with a measured normal risk ratio of 0.310.

At $\tau=0.005$ with ten calibration images, RALC also gained 3.085, 2.641 and 1.405 pp over ReCIRC-instance on B2, B0 and M2F WI, respectively, and incurred higher measured normal loss (supplement, Table~S17).

\paragraph{Matching mean normal loss.}
At the fixed budget, RALC incurred higher measured mean normal loss than Whole on every WI update, using 0.149--0.282 of the budget. To compare the methods at the same observed loss, we swept each rule's budget and interpolated its curve post hoc at a mean loss of 1,000 pixels per image on the 250 held-out normal images (Fig.~\ref{fig:riskbenefit}). RALC's mIoU on the 500 adverse images was then 1.135--2.428 pp higher than Whole's and 0.936--2.431 pp higher than ReCIRC-instance's (Table~\ref{tab:small}). All six paired intervals lay above zero when we resampled test images with the 50 calibration draws fixed (supplement, Table~S6).

The learned whole-image selector tests whether improved image ranking can recover the local gain. At the same post hoc loss target, RALC's adverse mIoU exceeded the learned selector's by 2.428, 1.336 and 0.945 pp on B2, B0 and M2F, respectively. All three paired intervals remained above zero when each replicate used a single calibration draw (supplement, Sec.~S12).

At each candidate level, the randomized control adopts the full update with a probability chosen to match the local candidate's expected replacement count on that image. At matched mean loss, $8{\times}8$ cells and RALC gained 0.880--1.945 and 0.956--2.001 pp over their respective randomized controls (Table~\ref{tab:small}; supplement, Sec.~S10).

\paragraph{Rule variants.}
Compared with the shared-bound variant, the default per-image rule retained more gain at the fixed budget and incurred more measured normal loss on all three WI updates. At matched loss the shared pixel rule exceeded the default per-image rule on B2 by 1.150 pp (paired 95\% interval 0.440--2.002). The B0 and M2F intervals included zero. Removing the optional image scorer also gave intervals spanning zero on all three adverse-scene comparisons at matched loss (supplement, Secs.~S5, S6 and S13).

\begin{figure*}[!t]
\centering\includegraphics[width=\textwidth]{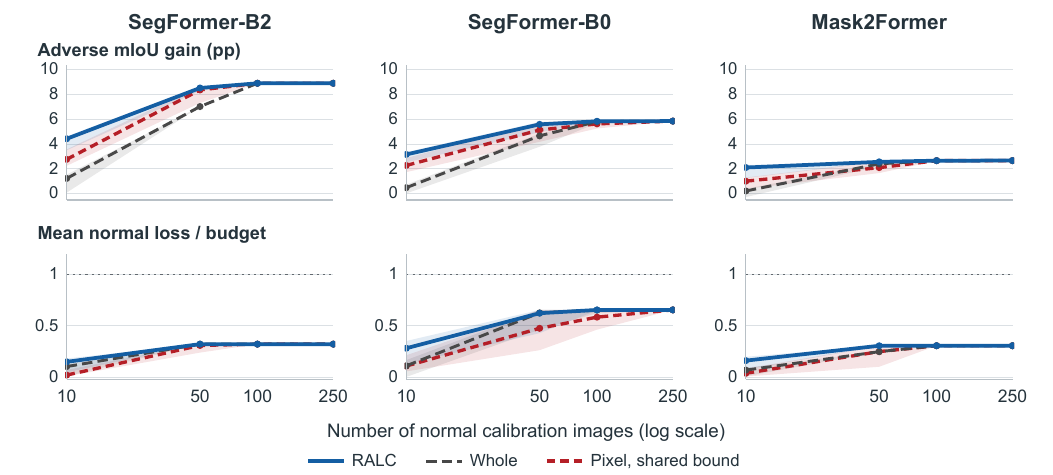}
\caption{Calibration size for WI updates on BDD at $\tau=0.005$. Top panels give adverse mIoU gain. Bottom panels give mean normal loss relative to budget (dotted line). Points average 50 calibration orders on the same test images. Bands span the 2.5th--97.5th percentiles. All orders use the full set at $n=250$.}
\label{fig:curves}
\end{figure*}

\subsection{Effect of calibration set size}
With no BDD calibration images ($n=0$), the budget caps the number of replacements on each image. With ten labeled normal images, the rule can adopt more changes on some images while keeping the expected loss on normal images within budget. At the same nominal budget, ten calibration images added 1.625--3.824 pp of adverse gain across the WI updates and 2.386 pp on B0 FT. Mean normal loss rose from 42--151 to 1,446--3,001 pixels/image across the four updates (supplement, Table~S19).

At the primary budget ($\tau=0.005$), Whole and RALC converge as the calibration set grows (Fig.~\ref{fig:curves}). At $n=250$, both accept all three WI updates in full. At the tighter prespecified budget ($\tau=0.0025$), full B0 WI exceeds the budget ($\rho=1.312$), while RALC retains 5.253 pp of adverse gain at $\rho=0.942$.

\begin{table}[t]
\caption{Adoption granularity on the original BDD test pool. Paired columns give adverse gain (pp) and normal risk ratio for the WI updates at the fixed budget in Table~\ref{tab:small}. Region-Conf.\ ranks 8-connected disagreement components by their mean confidence gap.}
\label{tab:local}
\centering
{\centering\small\setlength{\tabcolsep}{1.8pt}
\begin{tabular*}{\columnwidth}{@{\extracolsep{\fill}}lrrrrrr@{}}
\toprule
Controller & \multicolumn{2}{c}{B2} & \multicolumn{2}{c}{B0} & \multicolumn{2}{c}{M2F} \\
 & $\Delta_{\rm adv}$ & $\rho$ & $\Delta_{\rm adv}$ & $\rho$ & $\Delta_{\rm adv}$ & $\rho$ \\
\midrule
Whole & 1.235 & 0.103 & 0.492 & 0.110 & 0.225 & 0.068 \\
Region-Conf. & 2.533 & 0.123 & 1.690 & 0.220 & 1.364 & 0.139 \\
$32{\times}32$ grid & 4.182 & 0.145 & 3.054 & 0.291 & 2.012 & 0.161 \\
RALC & \textbf{4.427} & 0.149 & \textbf{3.166} & 0.282 & \textbf{2.114} & 0.161 \\
\bottomrule\end{tabular*}
}
\end{table}

\subsection{Effect of adoption granularity}
Keeping the calibration protocol fixed, we compare whole images, connected components, grids and pixels as adoption units. The largest connected component contains a median 40.7--51.3\% of disagreements across the three models. At fixed budget on B2, gain rose from 1.235 pp for Whole to 2.533 for components, 4.182 for $32{\times}32$ cells and 4.427 for pixels (Table~\ref{tab:local}; supplement, Fig.~S3).

On the same 500 adverse images, $8{\times}8$ cells recovered 93--98\% of RALC's advantage over Whole at matched mean normal loss (Fig.~\ref{fig:riskbenefit}, bars). Selecting individual pixels within cells added 0.058, 0.100 and 0.079 pp on B2, B0 and M2F. When we re-matched mean loss within each calibration draw, the pixel--grid intervals included zero on B2 and M2F (supplement, Table~S6). On the 399 adverse development images, cells containing both corrections and new errors accounted for 3.31--3.90\% of labeled disagreements (supplement, Sec.~S14).

\begin{figure*}[!t]
\centering\includegraphics[width=\textwidth]{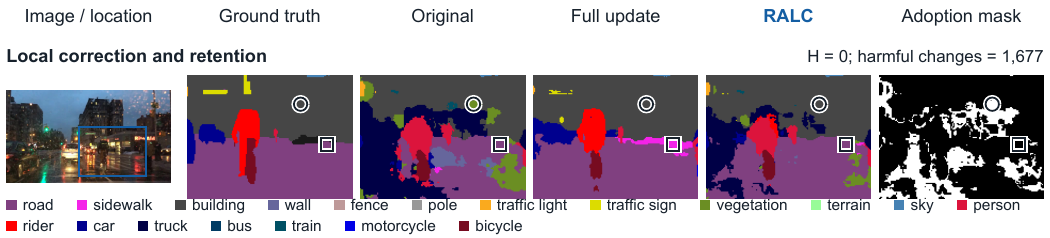}
\caption{Local adoption on the B2 WI adverse image with the largest net correction under the first BDD calibration draw. The circled change corrects an error; at the square, RALC retains a correct original label. The box on the image locates the enlarged region, and white mask pixels indicate adopted changes. Counts refer to the full image. Supplementary Sec.~S16 shows the complete images and normal-scene error cases.}
\label{fig:cases}
\end{figure*}

On the rainy image in Fig.~\ref{fig:cases}, RALC adopts the update's correction at the circled location and retains the correct original label at the square.

\begin{table*}[t]
\caption{Deployment on RTX 4080 SUPER. Each latency range (ms) spans the medians measured separately on the two input images. Candidate runs the updated model alone, Dual runs both, and Controller takes cached GPU predictions. Full RALC is timed end to end.}\label{tab:cost}
\centering\centering
\small
\setlength{\tabcolsep}{3pt}
\begin{tabular}{lrrrrrr}
\toprule
Model & Candidate & Dual & Full RALC & Controller & Full/cand. & Peak MiB \\
\midrule
B2 & 60.8--61.3 & 106.0--106.3 & 175.5--178.1 & 70.9--83.6 & 2.86--2.93 & 1822 \\
B0 & 32.4--32.7 & 50.7--51.0 & 116.4--121.1 & 70.7--73.3 & 3.56--3.73 & 589 \\
M2F & 122.2--130.8 & 230.8--232.3 & 298.4--302.9 & 68.6--93.2 & 2.32--2.44 & 1847 \\
\bottomrule
\end{tabular}

\end{table*}

\subsection{External evaluation on RaidaR}
\paragraph{Fixed budget.}
For RaidaR~\cite{raidar}, we kept the updates and scorers fixed and recalibrated the rules on ten sunny images. The 3,642 rainy test frames form two contiguous frame-ID blocks. At $\tau=0.005$, the learned whole-image selector accepted the full B2 update on 91.3\% of rainy images, gaining 1.673 pp against RALC's 1.711 pp. B0 gained 2.881 pp with RALC versus 0.182 pp with the learned selector, at higher sunny loss (4,093 vs.\ 1,938 pixels/image). For M2F, RALC gained 0.041 pp over the original; full adoption reduced rainy mIoU. Pixels added 0.018, 0.050 and 0.005 pp over $8{\times}8$ cells on B2, B0 and M2F.

\paragraph{Matching mean sunny loss.}
We matched mean sunny loss post hoc on RaidaR at 1,000 pixels/image. On B2, the learned whole-image selector's interpolated acceptance rate fell to 14.9\% and its gain to 0.070 pp, while RALC retained 1.912 pp through local changes. At 1,000 pixels/image, RALC led the learned selector by 2.031 pp on B0 and 0.295 pp on M2F. The B2 $8{\times}8$ grid gained 1.781 pp, capturing 93\% of RALC's advantage over the learned selector (supplement, Sec.~S12).

\subsection{Extension to multi-label classification}
On COCO2014~\cite{coco}, we adopt binary category labels rather than pixels. Four fixed updates use ResNet-50~\cite{resnet} and ConvNeXt-Tiny~\cite{convnext}, trained with ASL~\cite{asl} and adapted to synthetic low light and blur. Decision thresholds are set on separate development data before calibration. With ten clean calibration images and a budget of 0.25 label errors/image, RALC improves corrupted-image micro-F1 by 5.891--7.133 pp over the original classifiers, with measured mean clean loss of 0.136--0.147. Supplementary Sec.~S17 gives the protocol and per-update results.

\subsection{Normal-scene error analysis}
Across 200 splits of the 500 normal BDD images into calibration and test sets, B0 WI at $n=250$ and $\tau=0.0025$ had mean risk ratio 0.849; 24.0\% of splits had measured loss above budget (supplement, Sec.~S9).

In the normal B2 cases in supplementary Sec.~S16, RALC's merged prediction adds 42,324 net errors on one image; on another, corrections offset all 32,227 harmful changes. At $n=10$ and $\tau=0.005$, RALC changes an average of 10,935 correct labels per normal image. The measured risk ratio is $\rho=0.149$, and these flips affect 1.70\% of all pixels originally classified correctly on the normal test set. Recalibrating on harmful flips alone, RALC gained 1.447--3.001 pp over Whole and 0.127--0.254 pp over the $32{\times}32$ grid on the 500 adverse images. Its mean harmful-flip count on the 250 normal images used 49--53\% of the budget (supplement, Sec.~S3).

Under RALC, normal IoU fell below the original model for 1, 3 and 2 classes on B2, B0 and M2F, respectively. On B2, rider IoU changes from 35.47\% originally to 19.87\% under the full update and 21.30\% under RALC. Rider appears in 16 normal test images.

\subsection{Computational cost}
On two $720\times1280$ images, end-to-end RALC took 2.32--3.73 times as long as the updated model alone. With predictions cached on GPU, the controller took 69--93 ms (batch size one, five warm-ups, twenty repetitions, Table~\ref{tab:cost}).

\paragraph{Operating limits.}
A fixed pixel budget covers a smaller fraction of a higher-resolution image. Changing models or scoring rules requires recalibration. Shifts in the normal-image distribution can invalidate exchangeability~\cite{nonexchange,active}.

\section{Conclusion}
RALC turns deployment of a fixed segmentation update into a choice of which local changes the available calibration evidence can support. Its replacement-count bound ties the unknown test loss to the proposed changes, so reducing their scope can make an otherwise rejected update partly admissible. With the configuration fixed before calibration and exchangeable normal calibration and test images, the rule provides a finite-sample guarantee on expected normal-scene regression loss.

At matched measured normal loss, the segmentation experiments showed that within-image adoption retained gains that whole-image selectors left unused. Both small grids and pixels achieved this benefit. These results support local adoption when few calibration images restrict whole-image acceptance.

\section*{LLM Usage Statement}
Anthropic Claude and OpenAI Codex assisted with manuscript organization, drafting, translation, language editing, literature checks, and code preparation. OpenAI Codex also assisted with theoretical derivations and proof checks, experiment execution and analysis, plotting scripts, and the method diagram. Qwen provided feedback on the abstract and introduction, and OpenAI Codex assisted with pre-submission review. The author takes responsibility for the scientific claims, experimental results, figures, and references.
{\small\bibliographystyle{ieeenat_fullname}\bibliography{references}}

@inproceedings{wise,
  title = {{Robust Fine-Tuning of Zero-Shot Models}},
  author = {Wortsman, Mitchell and Ilharco, Gabriel and Kim, Jong Wook and Li, Mike and Kornblith, Simon and Roelofs, Rebecca and Lopes, Raphael Gontijo and Hajishirzi, Hannaneh and Farhadi, Ali and Namkoong, Hongseok and Schmidt, Ludwig},
  year = {2022},
  booktitle = {Proceedings of the IEEE/CVF Conference on Computer Vision and Pattern Recognition},
  pages = {7959--7971}
}

@inproceedings{pct,
  title = {{Positive-Congruent Training: Towards Regression-Free Model Updates}},
  author = {Yan, Sijie and Xiong, Yuanjun and Kundu, Kaustav and Yang, Shuo and Deng, Siqi and Wang, Meng and Xia, Wei and Soatto, Stefano},
  year = {2021},
  booktitle = {Proceedings of the IEEE/CVF Conference on Computer Vision and Pattern Recognition},
  pages = {14299--14308}
}

@inproceedings{mask2former,
  title = {{Masked-Attention Mask Transformer for Universal Image Segmentation}},
  author = {Cheng, Bowen and Misra, Ishan and Schwing, Alexander G. and Kirillov, Alexander and Girdhar, Rohit},
  year = {2022},
  booktitle = {Proceedings of the IEEE/CVF Conference on Computer Vision and Pattern Recognition},
  pages = {1290--1299}
}

@inproceedings{cityscapes,
  title = {{The Cityscapes Dataset for Semantic Urban Scene Understanding}},
  author = {Cordts, Marius and Omran, Mohamed and Ramos, Sebastian and Rehfeld, Timo and Enzweiler, Markus and Benenson, Rodrigo and Franke, Uwe and Roth, Stefan and Schiele, Bernt},
  year = {2016},
  booktitle = {Proceedings of the IEEE Conference on Computer Vision and Pattern Recognition},
  pages = {3213--3223}
}

@inproceedings{segformer,
  title = {{SegFormer: Simple and Efficient Design for Semantic Segmentation with Transformers}},
  author = {Xie, Enze and Wang, Wenhai and Yu, Zhiding and Anandkumar, Anima and Alvarez, Jose M. and Luo, Ping},
  year = {2021},
  booktitle = {Advances in Neural Information Processing Systems},
  volume = {34},
  pages = {12077--12090}
}

@inproceedings{acdc,
  title = {{ACDC: The Adverse Conditions Dataset With Correspondences for Semantic Driving Scene Understanding}},
  author = {Sakaridis, Christos and Dai, Dengxin and Van Gool, Luc},
  year = {2021},
  booktitle = {Proceedings of the IEEE/CVF International Conference on Computer Vision},
  pages = {10765--10775}
}

@inproceedings{muses,
  title = {{MUSES: The Multi-sensor Semantic Perception Dataset for Driving Under Uncertainty}},
  author = {Brödermann, Tim and Bruggemann, David and Sakaridis, Christos and Ta, Kevin and Liagouris, Odysseas and Corkill, Jason and Van Gool, Luc},
  year = {2024},
  doi = {10.1007/978-3-031-73202-7_2},
  url = {https://doi.org/10.1007/978-3-031-73202-7_2},
  booktitle = {Lecture Notes in Computer Science},
  pages = {21--38}
}

@inproceedings{bdd,
  title = {{BDD100K: A Diverse Driving Dataset for Heterogeneous Multitask Learning}},
  author = {Yu, Fisher and Chen, Haofeng and Wang, Xin and Xian, Wenqi and Chen, Yingying and Liu, Fangchen and Madhavan, Vashisht and Darrell, Trevor},
  year = {2020},
  booktitle = {Proceedings of the IEEE/CVF Conference on Computer Vision and Pattern Recognition},
  pages = {2636--2645}
}

@inproceedings{swin,
  title = {{Swin Transformer: Hierarchical Vision Transformer Using Shifted Windows}},
  author = {Liu, Ze and Lin, Yutong and Cao, Yue and Hu, Han and Wei, Yixuan and Zhang, Zheng and Lin, Stephen and Guo, Baining},
  year = {2021},
  booktitle = {Proceedings of the IEEE/CVF International Conference on Computer Vision},
  pages = {10012--10022}
}

@inproceedings{crc,
  title = {{Conformal Risk Control}},
  author = {Angelopoulos, Anastasios N. and Bates, Stephen and Fisch, Adam and Lei, Lihua and Schuster, Tal},
  year = {2024},
  url = {https://openreview.net/forum?id=33XGfHLtZg},
  booktitle = {International Conference on Learning Representations}
}

@article{ltt,
  title = {{Learn then test: Calibrating predictive algorithms to achieve risk control}},
  author = {Angelopoulos, Anastasios N. and Bates, Stephen and Candès, Emmanuel J. and Jordan, Michael I. and Lei, Lihua},
  year = {2025},
  doi = {10.1214/24-aoas1998},
  url = {https://doi.org/10.1214/24-aoas1998},
  journal = {The Annals of Applied Statistics},
  volume = {19},
  number = {2},
  pages = {1641--1662}
}

@article{rcps,
  title = {{Distribution-free, Risk-controlling Prediction Sets}},
  author = {Bates, Stephen and Angelopoulos, Anastasios and Lei, Lihua and Malik, Jitendra and Jordan, Michael},
  year = {2021},
  doi = {10.1145/3478535},
  url = {https://doi.org/10.1145/3478535},
  journal = {Journal of the ACM},
  volume = {68},
  number = {6},
  pages = {1--34}
}

@inproceedings{nonexchange,
  title = {{Non-Exchangeable Conformal Risk Control}},
  author = {Farinhas, António and Zerva, Chrysoula and Ulmer, Dennis and Martins, André F. T.},
  year = {2024},
  url = {https://arxiv.org/abs/2310.01262},
  booktitle = {International Conference on Learning Representations}
}

@article{wang,
  title = {{Should This Case Be Adapted? Prediction Fragmentation Controls Test-Time Adaptation}},
  author = {Wang, Lili and Li, Jing and Sun, Xiaowen and Hu, Xiangyu and Gu, Zhuangzhuang and Liu, Jian and Nelakuditi, Srihari and Tong, Yan},
  year = {2026},
  url = {https://arxiv.org/abs/2609.20700},
  journal = {arXiv preprint arXiv:2609.20700}
}

@inproceedings{active,
  title = {{Active, anytime-valid risk controlling prediction sets}},
  author = {Xu, Ziyu and Karampatziakis, Nikos and Mineiro, Paul},
  year = {2024},
  booktitle = {Advances in Neural Information Processing Systems},
  volume = {37},
  doi = {10.52202/079017-1920},
  pages = {60110--60132}
}

@inproceedings{tent,
  title = {{Tent: Fully Test-time Adaptation by Entropy Minimization}},
  author = {Wang, Dequan and Shelhamer, Evan and Liu, Shaoteng and Olshausen, Bruno and Darrell, Trevor},
  year = {2021},
  url = {https://github.com/DequanWang/tent},
  booktitle = {International Conference on Learning Representations}
}

@inproceedings{luo,
  title = {{Enhancing Image-Conditional Coverage in Segmentation: Adaptive Thresholding via Differentiable Miscoverage Loss}},
  author = {Luo, Rui and Bao, Jie and Su, Xiaoyi and Li, Wen and Cao, Suqun},
  year = {2026},
  booktitle = {International Conference on Learning Representations},
  pages = {73038--73065}
}

@inproceedings{cotta,
  title = {{Continual Test-Time Domain Adaptation}},
  author = {Wang, Qin and Fink, Olga and Van Gool, Luc and Dai, Dengxin},
  year = {2022},
  booktitle = {Proceedings of the IEEE/CVF Conference on Computer Vision and Pattern Recognition},
  pages = {7201--7211}
}

@inproceedings{eata,
  title = {{Efficient Test-Time Model Adaptation without Forgetting}},
  author = {Shuaicheng Niu and Jiaxiang Wu and Yifan Zhang and Yaofo Chen and Shijian Zheng and Peilin Zhao and Mingkui Tan},
  year = {2022},
  booktitle = {Proceedings of the 39th International Conference on Machine Learning},
  pages = {16888--16905},
  volume = {162},
  series = {Proceedings of Machine Learning Research},
  publisher = {PMLR}
}

@inproceedings{sar,
  title = {{Towards Stable Test-Time Adaptation in Dynamic Wild World}},
  author = {Niu, Shuaicheng and Wu, Jiaxiang and Zhang, Yifan and Wen, Zhiquan and Chen, Yaofo and Zhao, Peilin and Tan, Mingkui},
  year = {2023},
  url = {https://openreview.net/forum?id=g2YraF75Tj},
  booktitle = {International Conference on Learning Representations}
}

@inproceedings{selectivenet,
  title = {{SelectiveNet: A Deep Neural Network with an Integrated Reject Option}},
  author = {Yonatan Geifman and Ran El-Yaniv},
  year = {2019},
  booktitle = {Proceedings of the 36th International Conference on Machine Learning},
  pages = {2151--2159},
  volume = {97},
  series = {Proceedings of Machine Learning Research},
  publisher = {PMLR}
}

@inproceedings{soups,
  title = {{Model soups: averaging weights of multiple fine-tuned models improves accuracy without increasing inference time}},
  author = {Mitchell Wortsman and Gabriel Ilharco and Samir Ya Gadre and Rebecca Roelofs and Raphael Gontijo-Lopes and Ari S Morcos and Hongseok Namkoong and Ali Farhadi and Yair Carmon and Simon Kornblith and Ludwig Schmidt},
  year = {2022},
  booktitle = {Proceedings of the 39th International Conference on Machine Learning},
  pages = {23965--23998},
  volume = {162},
  series = {Proceedings of Machine Learning Research},
  publisher = {PMLR}
}

@inproceedings{distort,
  title = {{Fine-Tuning can Distort Pretrained Features and Underperform Out-of-Distribution}},
  author = {Kumar, Ananya and Raghunathan, Aditi and Jones, Robbie and Ma, Tengyu and Liang, Percy},
  year = {2022},
  booktitle = {International Conference on Learning Representations}
}

@misc{aacrc,
  author = {Blot, Vincent and Angelopoulos, Anastasios N. and Jordan, Michael I. and Brunel, Nicolas J.-B.},
  title = {Automatically Adaptive Conformal Risk Control},
  year = {2024},
  howpublished = {arXiv preprint arXiv:2406.17819},
  note = {Version 4, revised March 27, 2025},
  url = {https://arxiv.org/abs/2406.17819}
}

@misc{recirc,
  author = {{Marcondes e Resende}, Bruno and Graziadei, Helton and Ramos, Thiago Rodrigo and Izbicki, Rafael},
  title = {{ReCIRC}: Rectified Conformal Risk Control},
  year = {2026},
  howpublished = {arXiv preprint arXiv:2609.38112},
  url = {https://arxiv.org/abs/2609.38112}
}

@inproceedings{raidar,
  author={Jin, Jiongchao and Fatemi, Arezou and Lira, Wallace Michel Pinto and Yu, Fenggen and Leng, Biao and Ma, Rui and Mahdavi-Amiri, Ali and Zhang, Hao},
  title={{RaidaR}: A Rich Annotated Image Dataset of Rainy Street Scenes},
  booktitle={Proceedings of the IEEE/CVF International Conference on Computer Vision Workshops},
  pages={2951--2961},
  year={2021}
}

@inproceedings{trauble2021updates,
  author = {Tr{\"a}uble, Frederik and von K{\"u}gelgen, Julius and Kleindessner, Matth{\"a}us and Locatello, Francesco and Sch{\"o}lkopf, Bernhard and Gehler, Peter V.},
  title = {Backward-Compatible Prediction Updates: A Probabilistic Approach},
  booktitle = {Advances in Neural Information Processing Systems},
  volume = {34},
  year = {2021}
}

@inproceedings{coco,
  author = {Lin, Tsung-Yi and Maire, Michael and Belongie, Serge and Hays, James and Perona, Pietro and Ramanan, Deva and Doll{\'a}r, Piotr and Zitnick, C. Lawrence},
  title = {Microsoft {COCO}: Common Objects in Context},
  booktitle = {Computer Vision -- ECCV 2014},
  pages = {740--755},
  year = {2014},
  publisher = {Springer},
  doi = {10.1007/978-3-319-10602-1_48}
}

@inproceedings{resnet,
  author = {He, Kaiming and Zhang, Xiangyu and Ren, Shaoqing and Sun, Jian},
  title = {Deep Residual Learning for Image Recognition},
  booktitle = {Proceedings of the IEEE Conference on Computer Vision and Pattern Recognition},
  pages = {770--778},
  year = {2016}
}

@inproceedings{convnext,
  author = {Liu, Zhuang and Mao, Hanzi and Wu, Chao-Yuan and Feichtenhofer, Christoph and Darrell, Trevor and Xie, Saining},
  title = {A {ConvNet} for the 2020s},
  booktitle = {Proceedings of the IEEE/CVF Conference on Computer Vision and Pattern Recognition},
  pages = {11976--11986},
  year = {2022}
}

@inproceedings{asl,
  author = {Ridnik, Tal and Ben-Baruch, Emanuel and Zamir, Nadav and Noy, Asaf and Friedman, Itamar and Protter, Matan and Zelnik-Manor, Lihi},
  title = {Asymmetric Loss for Multi-Label Classification},
  booktitle = {Proceedings of the IEEE/CVF International Conference on Computer Vision},
  pages = {82--91},
  year = {2021}
}
\onecolumn
\providecommand{\WangRef}{24}

\setcounter{section}{0}
\renewcommand{\theHsection}{supp.\arabic{section}}
\setcounter{table}{0}
\renewcommand{\theHtable}{supp.\arabic{table}}
\setcounter{figure}{0}
\renewcommand{\theHfigure}{supp.\arabic{figure}}
\setcounter{equation}{0}
\renewcommand{\theHequation}{supp.\arabic{equation}}
\renewcommand{\thesection}{S\arabic{section}}
\renewcommand{\thetable}{S\arabic{table}}
\renewcommand{\thefigure}{S\arabic{figure}}

\begin{center}{\Large\bfseries\thetitle\par}\medskip{\large Supplementary Material\par}\end{center}
Throughout, RALC denotes the default pixel rule with the per-image replacement bound unless a different bound or granularity is specified. On the original BDD100K test pool, we report adverse-scene mIoU gain. On the separate official validation split, we report overall mIoU across scene conditions. Both comparisons measure normal loss on the original held-out normal images.

\section{Overall mIoU on the official validation split}
We ran all three frozen WI updates on the 1,000 images of the official BDD100K semantic segmentation validation split. We checked the file lengths and CRCs of all image--mask pairs against the Berkeley archive. Neither image identifiers nor SHA-256 hashes of compressed files or decoded RGB images matched the original 1,000-image pool. Two portrait images had landscape masks and no orientation tag, leaving 998 geometrically aligned pairs for scoring. The weather and time-of-day metadata used in the original pool contained no matching validation IDs. All reported validation mIoUs therefore pool the aligned images across scene conditions.

We kept the same image scorers, WI coefficients, 32 confidence bins and pixel budget ($\alpha\approx9{,}694$). Each of the 50 original calibration draws contained ten normal images, covering 225 distinct images across draws. For a grid rule, all disagreements in an origin-aligned cell share their mean confidence-gap score. Each granularity supplies its own calibration loss envelope. At a given level, let $B_x$ denote the number of proposed replacements on the test image. The per-image reserve is $B_x$, and the shared reserve is $\max(B_x,\max_{i\le n}B_i)$. These two rules share the same calibration loss sum within a granularity.

\begin{table}[ht]\centering
\caption{Utility is measured on the 998 aligned validation pairs. Gain is relative to the original segmentation model. Risk ratios and their 95\% test-bootstrap intervals use the original 250 held-out normal images. All three WI updates use $n=10$, $\tau=0.005$.}\label{tab:valrisk}
\begin{tabular}{llrrrr}
\toprule
Model & Rule & mIoU (\%) & Gain (pp) & $\overline H/\alpha$ & Risk interval \\
\midrule
SegFormer-B2 & Whole & 52.878 & 2.191 & 0.103 & [0.062, 0.147] \\
 & $32{\times}32$, per-image & 55.912 & 5.226 & 0.145 & [0.090, 0.204] \\
 & $8{\times}8$, shared & 53.423 & 2.737 & 0.020 & [0.008, 0.035] \\
 & $8{\times}8$, per-image & 55.944 & 5.258 & 0.147 & [0.092, 0.208] \\
 & Pixel, shared & 53.543 & 2.856 & 0.020 & [0.008, 0.034] \\
 & Pixel, per-image (RALC) & 55.977 & 5.291 & 0.149 & [0.092, 0.211] \\
\midrule
SegFormer-B0 & Whole & 45.267 & 1.096 & 0.110 & [0.071, 0.153] \\
 & $32{\times}32$, per-image & 48.038 & 3.867 & 0.291 & [0.205, 0.395] \\
 & $8{\times}8$, shared & 46.500 & 2.329 & 0.107 & [0.054, 0.177] \\
 & $8{\times}8$, per-image & 48.083 & 3.912 & 0.281 & [0.197, 0.382] \\
 & Pixel, shared & 46.574 & 2.403 & 0.108 & [0.055, 0.177] \\
 & Pixel, per-image (RALC) & 48.115 & 3.944 & 0.282 & [0.197, 0.383] \\
\midrule
Mask2Former & Whole & 56.619 & 0.193 & 0.068 & [0.044, 0.097] \\
 & $32{\times}32$, per-image & 57.920 & 1.494 & 0.161 & [0.110, 0.219] \\
 & $8{\times}8$, shared & 57.397 & 0.972 & 0.038 & [0.016, 0.071] \\
 & $8{\times}8$, per-image & 57.943 & 1.517 & 0.161 & [0.111, 0.218] \\
 & Pixel, shared & 57.452 & 1.026 & 0.037 & [0.015, 0.070] \\
 & Pixel, per-image (RALC) & 57.978 & 1.553 & 0.161 & [0.111, 0.216] \\
\bottomrule\end{tabular}\end{table}

For B2, B0 and M2F, respectively, overall validation mIoU was 50.686, 44.171 and 56.426\% before the update, and 57.750, 50.401 and 59.131\% under full WI adoption. The larger mIoUs under per-image bounds in Table~\ref{tab:valrisk} accompany higher measured risk than under shared bounds. Comparing pixels with $8{\times}8$ cells under the same per-image bound gives similar normal risk ratios for the two granularities on each model.

\clearpage
\section{Paired comparisons and deployment variability}
We resampled the same 998 validation image IDs for all rules and models in each of 2,000 bootstrap replicates. For each calibration draw we summed the resampled class counts, calculated mIoU over classes with nonzero union, and took the paired difference. The test-bootstrap interval uses the 2.5th and 97.5th percentiles of the mean difference across the 50 fixed draws. The single-deployment interval instead uses one common, uniformly selected draw per replicate. It describes variability across deployments and test samples. Image-ID resampling does not account for possible dependence within driving trips.

\begin{table}[ht]\centering
\caption{Paired differences in overall validation mIoU (pp). Grid denotes $8{\times}8$ cells unless otherwise indicated. Each $32{\times}32$ comparator uses the per-image bound.}\label{tab:valpaired}
\begin{tabular}{llrrr}
\toprule
Model & Contrast & Mean & Test bootstrap & Single deployment \\
\midrule
SegFormer-B2 & Pixel minus grid, per-image & 0.033 & [0.017, 0.049] & [-0.007, 0.102] \\
 & Pixel minus grid, shared & 0.120 & [0.099, 0.140] & [0.049, 0.439] \\
 & Per-image minus shared, pixel & 2.434 & [1.804, 3.079] & [1.041, 3.509] \\
 & Per-image minus shared, grid & 2.521 & [1.886, 3.159] & [1.088, 3.581] \\
 & RALC minus Whole & 3.099 & [2.342, 3.847] & [2.184, 4.135] \\
 & RALC minus $32{\times}32$ grid & 0.064 & [-0.007, 0.140] & [-0.079, 0.277] \\
\midrule
SegFormer-B0 & Pixel minus grid, per-image & 0.033 & [0.014, 0.051] & [-0.034, 0.104] \\
 & Pixel minus grid, shared & 0.074 & [0.061, 0.086] & [0.020, 0.295] \\
 & Per-image minus shared, pixel & 1.541 & [1.142, 1.941] & [0.275, 2.647] \\
 & Per-image minus shared, grid & 1.582 & [1.181, 1.985] & [0.271, 2.677] \\
 & RALC minus Whole & 2.848 & [2.316, 3.296] & [2.100, 3.697] \\
 & RALC minus $32{\times}32$ grid & 0.077 & [0.021, 0.128] & [-0.035, 0.230] \\
\midrule
Mask2Former & Pixel minus grid, per-image & 0.035 & [0.012, 0.054] & [-0.039, 0.111] \\
 & Pixel minus grid, shared & 0.054 & [0.038, 0.069] & [0.018, 0.126] \\
 & Per-image minus shared, pixel & 0.527 & [0.029, 1.001] & [-0.210, 1.360] \\
 & Per-image minus shared, grid & 0.545 & [0.053, 1.016] & [-0.171, 1.347] \\
 & RALC minus Whole & 1.360 & [0.822, 1.989] & [0.680, 2.192] \\
 & RALC minus $32{\times}32$ grid & 0.058 & [0.005, 0.106] & [-0.045, 0.200] \\
\bottomrule\end{tabular}\end{table}

The same calculations on the original 500 adverse images give Table~\ref{tab:original}. Its calibration column varies the draw with test images held fixed. These original comparisons use the positive net loss $H$ and the $32{\times}32$ grid.

\begin{table}[ht]\centering
\caption{Original adverse test set. RALC minus Whole or the $32{\times}32$ grid in adverse mIoU (pp) under calibration on $H$. Joint variation uses one calibration draw per resampled test set.}\label{tab:original}
\centering\small\setlength{\tabcolsep}{4pt}\renewcommand{\arraystretch}{1.15}
\begin{tabular}{llrrrr}
\toprule
Model & Comparator & $\Delta$ (pp) & Calibration & Test bootstrap & Joint variation \\
\midrule
B2 & Whole & 3.192 & [2.955, 3.698] & [2.770, 3.831] & [2.669, 3.955] \\
B2 & Grid-Conf. & 0.245 & [0.181, 0.418] & [0.156, 0.360] & [0.116, 0.452] \\
B0 & Whole & 2.675 & [2.422, 2.969] & [2.313, 3.211] & [2.251, 3.310] \\
B0 & Grid-Conf. & 0.113 & [0.067, 0.182] & [0.052, 0.168] & [0.021, 0.209] \\
M2F & Whole & 1.889 & [1.260, 2.268] & [1.164, 2.894] & [1.035, 3.010] \\
M2F & Grid-Conf. & 0.102 & [0.040, 0.183] & [0.057, 0.160] & [-0.012, 0.242] \\
\bottomrule\end{tabular}

\end{table}

\clearpage
\section{Recalibration with no credit for corrections}
For the adopted set $A$, define $G(A)=\sum_{p\in A\cap V}\mathbf{1}\{\hat y^0_p=y_p,\ \hat y^1_p\ne y_p\}$, where $V$ is the set of valid labeled pixels. This loss counts every newly introduced error. It decreases as the nested sets shrink and satisfies $G(A)\le |A|$. Each rule can therefore use the original change-count bound and substitute $G$ for its calibration loss. We recalibrated Whole, the $32{\times}32$ grid and RALC separately on the original normal draws, reusing the frozen predictions. All three use the per-image reserve.

\begin{table}[ht]\centering
\caption{Direct calibration on harmful replacements. Utility uses the original 500 adverse images and risk uses the original 250 normal test images. The budget and 50 calibration draws are the same as in the net-loss experiment.}\label{tab:g}
\begin{tabular}{llrrrr}
\toprule
Model & Rule & Adverse mIoU (\%) & Gain (pp) & Mean $G$ (pixels) & $\overline G/\alpha$ \\
\midrule
SegFormer-B2 & Whole & 32.696 & 0.150 & 1,397.9 & 0.144 \\
 & $32{\times}32$ grid & 35.442 & 2.897 & 4,689.6 & 0.484 \\
 & RALC & 35.696 & 3.151 & 4,762.1 & 0.491 \\
\midrule
SegFormer-B0 & Whole & 28.850 & 0.040 & 390.8 & 0.040 \\
 & $32{\times}32$ grid & 30.995 & 2.185 & 4,857.7 & 0.501 \\
 & RALC & 31.137 & 2.327 & 5,111.1 & 0.527 \\
\midrule
Mask2Former & Whole & 40.993 & 0.036 & 1,043.4 & 0.108 \\
 & $32{\times}32$ grid & 42.312 & 1.356 & 4,839.6 & 0.499 \\
 & RALC & 42.439 & 1.483 & 4,754.7 & 0.490 \\
\bottomrule\end{tabular}\end{table}

\begin{table}[ht]\centering
\caption{RALC minus comparator in adverse mIoU (pp) after recalibration on $G$. Intervals use the paired procedures of Section~S2 on the original 500 adverse images.}\label{tab:gdelta}
\begin{tabular}{llrrr}
\toprule
Model & Comparator & Mean & Test bootstrap & Single deployment \\
\midrule
SegFormer-B2 & Whole & 3.001 & [2.663, 3.538] & [2.333, 3.861] \\
 & $32{\times}32$ grid & 0.254 & [0.186, 0.358] & [0.157, 0.386] \\
\midrule
SegFormer-B0 & Whole & 2.287 & [2.009, 2.724] & [1.858, 2.903] \\
 & $32{\times}32$ grid & 0.142 & [0.081, 0.203] & [0.042, 0.237] \\
\midrule
Mask2Former & Whole & 1.447 & [0.919, 2.043] & [0.545, 2.558] \\
 & $32{\times}32$ grid & 0.127 & [0.081, 0.184] & [-0.022, 0.251] \\
\bottomrule\end{tabular}\end{table}

\section{CPU timing from cached disagreements and confidence gaps}
Starting from cached disagreement coordinates and confidence differences, we timed binning, budget selection and binary mask construction on one CPU process. We used the first 32 validation images in filename order and the first calibration draw. Each reported time is the median over images of three-run medians after one warm-up. With the per-image bound, the pixel controller took 0.85, 1.15 and 0.91\,ms on B2, B0 and M2F, respectively. The $8{\times}8$ controller took 1.55, 2.12 and 1.73\,ms. Grid scoring includes the calculation of cell means. The end-to-end timings in the main paper include both segmentation networks and output transfer. They exclude disk I/O, initialization and offline calibration.
\clearpage
\section{Utility at matched measured normal risk}
We used the three frozen WI updates and the same 50 draws of ten normal calibration images as in the main comparisons. For each rule we evaluated 77 nominal budgets: $\tau$ from 0 to 0.018 in steps of 0.00025, followed by 0.020, 0.025, 0.030 and 0.040. Each budget gives one point on a curve. Its horizontal coordinate is the mean positive net increase in errors ($\overline H$, pixels per image) on the original 250 held-out normal images. Its vertical coordinate is the mIoU gain over the original segmenter on either the 500 original adverse images or the 998 aligned official validation images. We pooled class counts within each draw before computing mIoU, then averaged over the 50 draws.

To compare rules at $\overline H=1{,}000$, we followed each curve in order of increasing nominal budget and linearly interpolated between the two adjacent points that first bracketed the target. For example, B2 RALC had mean losses of 947.1 and 1,054.1 pixels at $\tau=0.00375$ and 0.004, with adverse gains of 3.552 and 3.783 pp. Interpolation gave 3.666 pp at 1,000 pixels. B2 Whole reached the same measured risk at an interpolated gain of 1.238 pp. This comparison uses held-out normal labels after evaluation to align operating points. Deployment uses the budget fixed before calibration. All four targets from 500 to 2,000 pixels in Table~\ref{tab:sensitivity} lie within the measured range of every rule and model compared there.

For the paired 95\% intervals, we resampled the normal image IDs and, independently, the utility evaluation image IDs with replacement. Within each of 1,000 replicates, all rules shared both samples and the original 50 calibration draws. We recomputed each risk curve, matched it at 1,000 pixels, and took the difference in interpolated gains. Tables~\ref{tab:matchedadverse} and~\ref{tab:matchedofficial} give the 2.5th and 97.5th percentiles. These intervals condition on the 50 calibration draws. For the pixel and $8{\times}8$ grid comparison in Table~\ref{tab:matchedadverse}, we also selected one calibration draw uniformly per replicate and used it for both rules. We evaluated that draw at operating points matched either within the draw or across all 50 draws. All replicates bracketed the target.

\begin{table}[ht]\centering
\caption{Original 500 adverse BDD100K images at $\overline H=1{,}000$ pixels per normal image. The upper rows give mean paired differences in adverse mIoU (pp) with 95\% intervals. Both grids and RALC use the per-image bound. The last two rows give single-draw 95\% intervals for RALC minus the $8{\times}8$ grid.}
\label{tab:matchedadverse}
{\small\setlength{\tabcolsep}{4pt}\begin{tabular}{lccc}
\toprule
Contrast & SegFormer-B2 & SegFormer-B0 & Mask2Former \\
\midrule
RALC minus Whole & \(2.428~[1.796, 3.105]\) & \(1.414~[0.936, 1.927]\) & \(1.135~[0.656, 1.817]\) \\
RALC minus ReCIRC-instance & \(2.431~[1.882, 3.112]\) & \(1.422~[0.984, 1.965]\) & \(0.936~[0.317, 1.761]\) \\
RALC minus $32{\times}32$ grid & \(0.179~[0.077, 0.334]\) & \(0.285~[0.153, 0.422]\) & \(0.232~[0.082, 0.388]\) \\
RALC minus $8{\times}8$ grid & \(0.058~[0.016, 0.122]\) & \(0.100~[0.043, 0.148]\) & \(0.079~[0.024, 0.222]\) \\
Pixel: per-image minus shared & \(-1.150~[-2.002, -0.440]\) & \(-0.362~[-0.751, 0.016]\) & \(0.065~[-0.411, 0.853]\) \\
\midrule Single draw, within-draw match & \([-0.038, 0.154]\) & \([0.024, 0.204]\) & \([-0.037, 0.295]\) \\
Single draw, pooled match & \([0.003, 0.142]\) & \([0.035, 0.162]\) & \([-0.001, 0.283]\) \\
\bottomrule\end{tabular}
}
\end{table}

\begin{table}[ht]\centering
\caption{Official validation split. Paired differences in \emph{overall} mIoU (pp) on the 998 aligned images at the same measured normal risk, with 95\% intervals. The pixel--grid comparison gives both rules the per-image bound.}
\label{tab:matchedofficial}
{\small\setlength{\tabcolsep}{4pt}\begin{tabular}{lccc}
\toprule
Contrast & SegFormer-B2 & SegFormer-B0 & Mask2Former \\
\midrule
RALC minus Whole & \(2.626~[1.487, 3.150]\) & \(1.550~[0.999, 2.091]\) & \(0.977~[0.667, 1.465]\) \\
RALC minus $8{\times}8$ grid & \(0.056~[-0.0005, 0.103]\) & \(0.075~[0.032, 0.155]\) & \(0.072~[0.029, 0.113]\) \\
Pixel: per-image minus shared & \(0.117~[-0.585, 0.674]\) & \(0.228~[-0.234, 0.773]\) & \(-0.019~[-0.452, 0.578]\) \\
$8{\times}8$ grid: per-image minus shared & \(0.132~[-0.575, 0.707]\) & \(0.223~[-0.261, 0.755]\) & \(-0.022~[-0.458, 0.570]\) \\
\bottomrule\end{tabular}
}
\end{table}

For the B2 and B0 pixel families on the original adverse images, the shared bound gave a higher interpolated gain at 1,000 pixels. The paired interval excluded zero only for B2. On official validation images at this risk, each paired interval for the per-image minus shared bound included zero across all three architectures with $8{\times}8$ cells (Table~\ref{tab:matchedofficial}).

\clearpage
\section{Image-scorer ablation}
The ablated rule uses the common cutoffs $b=0,\ldots,32$. At cutoff $b$ it replaces the disagreeing labels with confidence-gap bin $k_p\ge b$. It computes its own calibration loss envelope on this family and uses the same per-image replacement-count bound as RALC. The selected cutoff can still vary across test images because their replacement counts differ.

\paragraph{Fitting the image scorer.}
The scorer uses model entropy and confidence, the difference in mean entropy, the fraction of disagreeing pixels, and image brightness, contrast, dark-channel and color statistics. We compute mean entropy and confidence on the native logit grids and RGB statistics and disagreement on the input grid. For pixel confidence gaps we upsample the logits before applying softmax. For each fixed update, standardize the features using their training means and population standard deviations plus $10^{-9}$ in the denominator. Add an intercept and fit logistic regression from zero weights with 50 Newton steps and an $\ell_2$ penalty of $0.01$ on all coefficients, including the intercept (summed loss). Clip the logit to $[-30,30]$ before the sigmoid.

\begin{figure}[ht]\centering
\includegraphics[width=.96\linewidth]{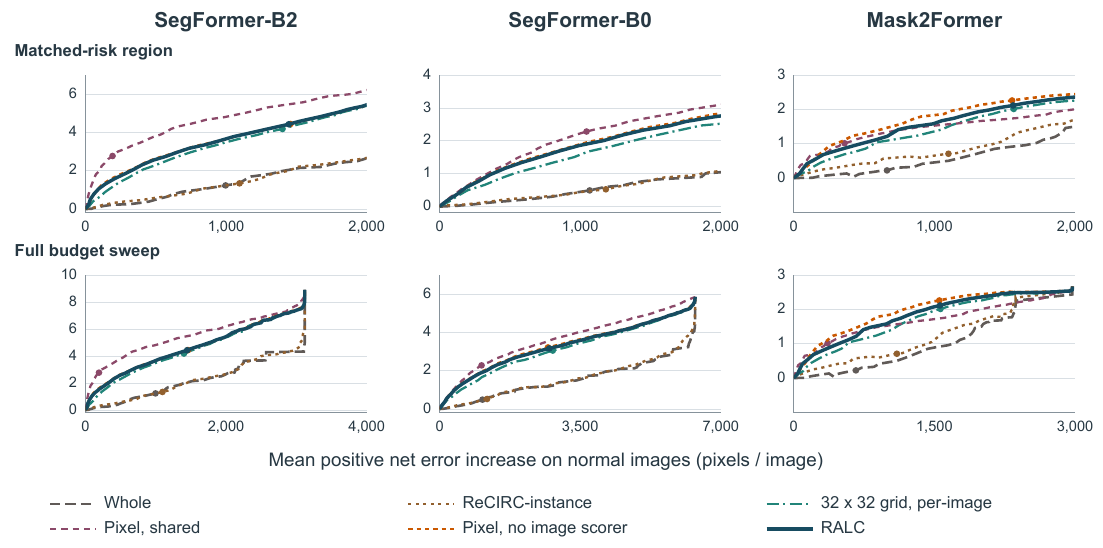}
\caption{Risk--benefit curves for the original 500 adverse images. The vertical axis gives adverse mIoU gain over the original model in pp. In each column the top panel shows the matched-risk region and the bottom panel the full sweep. Lines connect adjacent nominal budgets; dots mark $\tau=0.005$ when in range. The $32{\times}32$ grid, scorer-free pixel rule, ReCIRC-instance and RALC all use the per-image bound.}
\label{fig:matchedadverse}
\end{figure}

\begin{table}[ht]\centering
\caption{Removing the image scorer. Positive differences favor the rule with common confidence-gap cutoffs. Entries are paired mIoU differences in pp and 95\% intervals. At fixed budget each bootstrap replicate uses $\tau=0.005$ for both rules; at matched risk it repeats the interpolation.}
\label{tab:scorer}
{\small\setlength{\tabcolsep}{4pt}\begin{tabular}{llccc}
\toprule
Evaluation images & Comparison & SegFormer-B2 & SegFormer-B0 & Mask2Former \\
\midrule
Original adverse & $\tau=0.005$ & \(0.015~[-0.015, 0.049]\) & \(0.051~[0.013, 0.100]\) & \(0.145~[0.046, 0.298]\) \\
Original adverse & $\overline H=1{,}000$ & \(-0.013~[-0.117, 0.138]\) & \(0.032~[-0.046, 0.116]\) & \(0.241~[-0.045, 0.575]\) \\
Official overall & $\tau=0.005$ & \(0.044~[0.016, 0.071]\) & \(0.178~[0.114, 0.243]\) & \(0.053~[0.007, 0.102]\) \\
Official overall & $\overline H=1{,}000$ & \(0.048~[-0.051, 0.163]\) & \(0.312~[0.159, 0.426]\) & \(0.169~[-0.012, 0.316]\) \\
\bottomrule\end{tabular}
}
\end{table}

At $\tau=0.005$, removing the scorer changed the mean normal loss from 1,446.3 to 1,458.4 pixels for B2, from 2,730.4 to 2,697.8 for B0, and from 1,559.9 to 1,552.7 for M2F. At matched risk, B0 gained 0.312 pp in overall validation mIoU without the scorer. On the original adverse images, all three matched-risk intervals included zero.

\clearpage
\section{Validation curves and sensitivity to the target risk}
\begin{figure}[ht]\centering
\includegraphics[width=.96\linewidth]{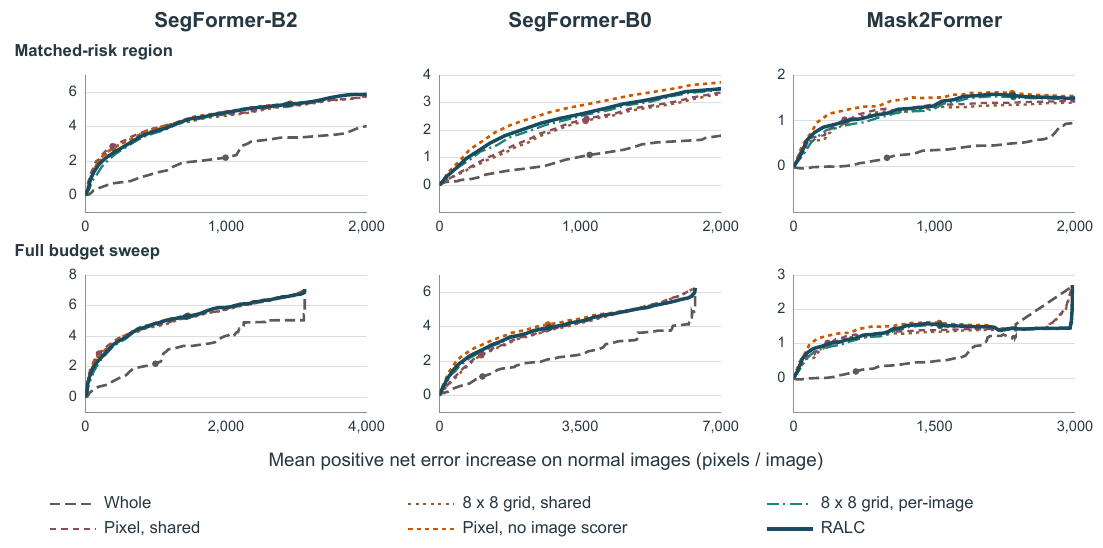}
\caption{Risk--benefit curves for the 998 official validation images, pooled across scene conditions. The vertical axis gives overall mIoU gain in pp. The horizontal axis uses the same 250 normal test images as Fig.~\ref{fig:matchedadverse}. Dots mark $\tau=0.005$. Solid and dashed curves can overlap where the rules have similar utility and risk.}
\label{fig:matchedofficial}
\end{figure}

\begin{table}[ht]\centering
\caption{Sensitivity on the original adverse test images. Entries are RALC minus the comparator in adverse mIoU (pp), interpolated separately at each target. Table~\ref{tab:matchedadverse} gives paired intervals at 1,000 pixels per image.}
\label{tab:sensitivity}
{\small\begin{tabular}{rlrrr}
\toprule
Mean $H$ (pixels) & Comparator & SegFormer-B2 & SegFormer-B0 & Mask2Former \\
\midrule
500 & Whole & 2.099 & 1.042 & 0.883 \\
500 & ReCIRC-instance & 2.005 & 0.986 & 0.600 \\
1,000 & Whole & 2.428 & 1.414 & 1.135 \\
1,000 & ReCIRC-instance & 2.431 & 1.422 & 0.936 \\
1,500 & Whole & 2.422 & 1.561 & 1.185 \\
1,500 & ReCIRC-instance & 2.373 & 1.566 & 0.960 \\
2,000 & Whole & 2.785 & 1.709 & 0.860 \\
2,000 & ReCIRC-instance & 2.774 & 1.670 & 0.666 \\
\bottomrule\end{tabular}
}
\end{table}

For the original adverse images, Fig.~\ref{fig:matchedadverse} shows the $32{\times}32$ grid curve and Table~\ref{tab:matchedadverse} reports the $8{\times}8$ grid comparison at matched risk. Figure~\ref{fig:matchedofficial} shows the $8{\times}8$ grid curve on the separate official validation images.

\clearpage
\section{Reproducing the ReCIRC-instance comparison}
ReCIRC-instance adapts learned risk-curve inversion to the frozen WI segmentation updates. For each update we fit one regressor before BDD calibration. It predicts a loss curve for each image and uses that curve to index the pixel candidates by a common predicted error count. Labeled normal calibration images then determine how much of this family can be adopted.

\paragraph{Training examples and target.}
We used the 300 normal images in the image-scorer training pool (200 Cityscapes and 100 ACDC). For each image $x_i$ and each of the 129 levels $\lambda=-32,\ldots,96$, we computed the original scorer's bin cutoff $b_i(\lambda)$ from Eq.~\eqref{eq:bin-threshold}. Write $H_i(b)$ for the positive net error increase when pixels in confidence-gap bins $b,\ldots,31$ are replaced; $b=32$ restores the original labels. The regression target at level $\lambda$ was
\[
  \frac{1}{P_i}\max_{b_i(\lambda)\le b\le32} H_i(b),
\]
where $P_i$ is the full input image area in pixels. Normalizing by area makes training and query images comparable across resolutions.

Each regression example concatenated the normalized level $(\lambda+32)/128$ with ten image features: mean predictive entropy and confidence for each of the two models, the change in mean entropy, the fraction of pixels with disagreeing labels, and four RGB statistics. With channels in $[0,1]$, the latter were the mean and standard deviation of $(R+G+B)/3$, the mean of the $15{\times}15$ dark channel, and $\sqrt{\operatorname{Var}(R-G)+\operatorname{Var}((R+G)/2-B)}$. Each image thus contributed 129 training examples; all levels of an image stayed in the same fold.

\paragraph{Fit and model selection.}
We used scikit-learn 1.7.2's histogram gradient boosting regressor with squared-error loss, 150 iterations, learning rate 0.05, at least 30 examples per leaf, early stopping disabled, and random seed 20261020. Three-fold GroupKFold kept images from a common recording together (217 groups). For Cityscapes, the group key combined the city and sequence ID; for ACDC, it used the filename prefix preceding the frame ID. We compared $(\text{maximum leaves},\,\ell_2\text{ penalty})$ settings $(15,1)$, $(31,1)$, and $(15,10)$. We averaged validation squared error first over levels within each image, then over all held-out training images. The lowest-error settings were $(15,1)$ for B2 and B0 and $(15,10)$ for M2F. We refit each selected regressor on all 300 images.

\paragraph{Inverting the learned curve.}
On a BDD image, we clipped the 129 predictions to $[0,1]$, multiplied by the image area $720{\times}1280=921{,}600$, and fitted a nonincreasing isotonic curve over $\lambda$. We set its final value to zero. The internal risk grid contains zero plus 256 geometrically spaced values from 1 to 921,600 inclusive. For each positive grid value $a$, we take the \emph{first} level whose fitted loss is at most $a$, then use its original pixel cutoff $b_x(\lambda)$ to form $A_x(a)$. At $a=0$ we explicitly use the empty set. Increasing $a$ can only enlarge the candidate. The grid is fixed across images and distinct from the deployment budget $\alpha$.

\paragraph{Offline calibration and online choice.}
For each labeled calibration image we recomputed the observed loss on the inverted candidates. Because $a$ runs from conservative to permissive, its envelope is the running maximum
\[
 E_i(a)=\max_{a'\le a}H_i\bigl(A_i(a')\bigr).
\]
ReCIRC-instance stores $\sum_{i=1}^n E_i(a)$ at each grid point. On the test image it takes the \emph{largest} $a$ for which $\sum_{i=1}^n E_i(a)+|A_x(a)|\le(n+1)\alpha$. The regressor maps the common grid to image-specific cutoffs. The stored observed losses and the current replacement count determine feasibility. ReCIRC-shared uses $\max(|A_x(a)|,\max_i|A_i(a)|)$ in the same check. ReCIRC-global reserves 921,600 pixels at positive grid values and zero at rollback. All three variants reuse the same inverted family, frozen predictions and calibration draws.

\clearpage
\section{Sensitivity to the choice of calibration images}
\paragraph{Reconstructing the original BDD pool and draws.}
Intersect the filename stems of non-test images, train-ID masks (strip \emph{\_train\_id}) and legacy weather-metadata JSON files. Among the 3,430 matches, 1,913 dry-day and 642 adverse images meet the conditions in the main text. Sort each list by filename stem. Using one NumPy PCG64 generator seeded with 12102025, permute the normal list first and the adverse list second and take the first 500 of each. The selected stems correspond to distinct videos. With a fresh generator seeded with 12102026, permute the sorted normal stems and use the first 250 for calibration. Sort those 250 again, then use a fresh generator seeded with 12102027 to permute that sorted list 50 times in succession. Each calibration size uses a prefix of the corresponding permutation.

\begin{figure}[ht]\centering
\includegraphics[width=.96\linewidth]{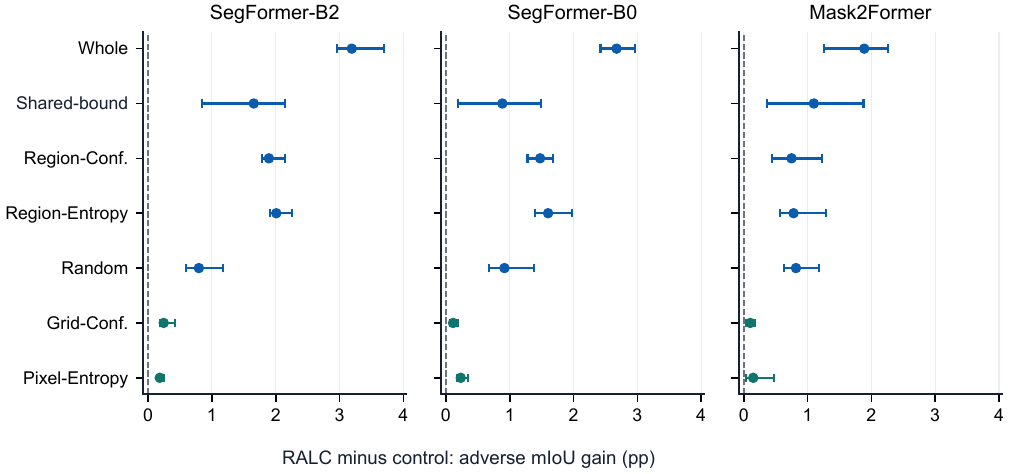}
\caption{RALC minus each local or whole-image control on the original adverse BDD test images. Grid-Conf.\ uses $32{\times}32$-pixel cells. Points are mean paired differences in adverse mIoU gain across the 50 draws of ten normal calibration images at $\tau=0.005$. Whiskers show the 2.5th and 97.5th percentiles across draws with the test images held fixed.}
\label{fig:ordering-draws}
\end{figure}

Figure~\ref{fig:ordering-draws} holds the original BDD test images fixed and varies the ten normal calibration images. Table~\ref{tab:original} also resamples the adverse test images. These two summaries answer different questions about how much of the update is retained.

\paragraph{Reallocating the normal pool.}
Across 72 prespecified settings on the fixed BDD split, the largest RALC mean risk ratio is 0.988. We sorted the 500 normal images by filename and generated 200 permutations of that list with a single NumPy PCG64 generator (seed 12102034). In each permutation, the first 250 images formed the calibration pool and the last 250 formed the normal test set. At each calibration size we used the corresponding prefix of the calibration pool. The highest allocation-mean ratio is 0.928 over 144 settings covering Whole, Shared-bound and RALC. For RALC on B0 at $\tau=0.0025$ and $n=250$, the FT and WI means are 0.820 and 0.849, although 21.0\% and 24.0\% of allocations exceed the allowance. This variation matters to a user who deploys a single calibration draw.

\paragraph{Development pools and recording-based calibration.}
At $n=10$ and $\tau=0.005$, the WI results on Cityscapes/ACDC retain 4.463--6.359 pp more adverse gain under RALC than under Whole. The corresponding range on MUSES is 1.984--3.597 pp. Under $\tau=0.0025$ with half-split calibration, the two ranges are 1.749--3.147 and 3.496--5.423 pp. These pools informed the design of the method. The BDD evaluation applies the resulting fixed rules to held-out scenes.

On MUSES, we also grouped consecutive frames by recording before splitting. Consecutive frames from one recording may share weather, illumination and road layout. The 350 normal MUSES images cover only 19 eligible recordings. Under recording-based evaluation, MUSES offers only 5--14 calibration recordings after splitting. The three MUSES WI updates retain 2.518--3.506 pp of adverse gain under RALC in that evaluation. A larger frame count from the same recordings does not increase this calibration sample size.

\clearpage
\section{Randomized whole-image adoption}
We paired each local family (pixels or $8{\times}8$ cells, both using the per-image bound) with a whole-image rule that can vary how often it accepts an update. All comparisons in this section use the three WI updates, the original BDD normal and adverse test sets, and the same 50 draws of ten calibration images as in Sec.~S5.

\paragraph{Control and calibration.}
For an image $x$ and level $\lambda$, let $D(x)$ be the full disagreement set and $A(x,\lambda)$ the candidate in the paired local family. Let $\pi(x,\lambda)=|A(x,\lambda)|/|D(x)|$, with $\pi=0$ if the models agree everywhere. The control uses the full update with probability $\pi$ and the original prediction otherwise. At this level its expected replacement count is $\pi|D|=|A|$. If $H_{\rm full}(x,y)$ is the positive part of the full update's net error increase, the control's expected loss is $\pi H_{\rm full}\leq\pi|D|=|A|$. A realized full adoption replaces all $|D|$ disagreeing labels.

For each level, sum $\pi(x_i,\lambda)H_{{\rm full},i}$ over the ten labeled calibration images. These losses are nonincreasing as the nested sets shrink. For a test image choose the largest candidate satisfying
\[
 \sum_{i=1}^{n}\pi(x_i,\lambda)H_{{\rm full},i}+|A(x,\lambda)|\ \le\ (n+1)\alpha .
\]
The local family uses its own calibrated losses. The two rules have equal expected counts at a common level and can select different levels after calibration. Both use the frozen image scores and candidate levels.

\paragraph{Risk matching and adverse utility.}
We used the 77 budgets and first-crossing interpolation in Sec.~S5. Normal risk averages $\pi H_{\rm full}$ at the selected level over 250 held-out normal images and the 50 draws. For adverse mIoU, we sampled the whole-image decisions on the 500 adverse images, pooled class counts for each realization and then averaged mIoU over 512 realizations per draw (PCG64 seed 33427). Within a draw, we used the same random numbers at adjacent budgets. Interpolation aligns each rule at a measured mean normal loss of 1,000 pixels after evaluation. Table~\ref{tab:randomwhole} also reports the fixed-budget results.

\begin{table}[ht]\centering
\caption{Local adoption and its paired randomized whole-image control (Rand.) on the original BDD test pool. Gains are adverse mIoU points over the original model. At $\tau=0.005$, $\overline H$ is the mean loss in pixels per held-out normal image. At matched loss, $\Delta$ is local minus Rand., calculated before rounding. The final column gives the single-calibration interval for that difference.}
\label{tab:randomwhole}
\small\setlength{\tabcolsep}{5pt}
\begin{tabular}{llrrrrrrrr}\toprule
 & & \multicolumn{4}{c}{Fixed budget ($\tau=0.005$)} & \multicolumn{4}{c}{Matched loss ($\overline H=1{,}000$)}\\
\cmidrule(lr){3-6}\cmidrule(lr){7-10}
 & & \multicolumn{2}{c}{Gain (pp)} & \multicolumn{2}{c}{$\overline H$ (pixels)} & \multicolumn{3}{c}{Gain or difference (pp)} & \\
Model & Family & Local & Rand. & Local & Rand. & Local & Rand. & $\Delta$ & Single-draw 95\% interval\\
\midrule
SegFormer-B2 & Pixels & 4.427 & 3.551 & 1,446 & 1,984 & 3.666 & 1.666 & 2.001 & [0.745, 3.556] \\
 & $8{\times}8$ cells & 4.356 & 3.552 & 1,426 & 1,987 & 3.608 & 1.663 & 1.945 & [0.683, 3.463] \\
\addlinespace
SegFormer-B0 & Pixels & 3.166 & 2.248 & 2,730 & 3,483 & 1.862 & 0.613 & 1.249 & [0.609, 2.064] \\
 & $8{\times}8$ cells & 3.132 & 2.246 & 2,725 & 3,475 & 1.762 & 0.606 & 1.156 & [0.492, 2.011] \\
\addlinespace
Mask2Former & Pixels & 2.114 & 1.276 & 1,560 & 1,729 & 1.592 & 0.636 & 0.956 & [0.206, 2.140] \\
 & $8{\times}8$ cells & 2.081 & 1.271 & 1,562 & 1,726 & 1.513 & 0.633 & 0.880 & [0.134, 1.977] \\
\bottomrule\end{tabular}
\end{table}

\paragraph{Single-calibration variability.}
Each of 1,000 replicates picks one of the 50 calibration draws uniformly and resamples the normal and adverse image IDs separately with replacement. The two rules share the draw and both sets of image IDs. Each sampled occurrence receives an independent uniform variate for the randomized control, reused across all budgets. We use realized normal losses and pooled adverse class counts to reconstruct both curves and re-match within the replicate (PCG64 seed 33428). All replicates bracketed the target. The interval spans the 2.5th and 97.5th percentiles of the paired differences and includes variation across calibration draws, test samples and randomized decisions.

\clearpage
\section{Model updates and weight interpolation}
\paragraph{Initial models and training split.}
The initial checkpoints are the Cityscapes releases \textsf{nvidia/segformer-b2-finetuned-cityscapes-1024-1024}, \textsf{nvidia/segformer-b0-finetuned-cityscapes-1024-1024}, and \textsf{facebook/mask2former-swin-base-IN21k-cityscapes-semantic}. The ACDC adverse training sequences are split within each condition into two groups by assigning sequences in descending frame-count order to the group with fewer frames, breaking ties by sequence name. The 807 images in group A train the update; the adverse images in the scorer pool come from group B. For each architecture we initialize from its Cityscapes checkpoint and fine-tune for 10,000 iterations at batch size two.

\paragraph{Mask2Former checkpoint loading.}
With Transformers 4.46.3, the checkpoint uses older Swin backbone key names. When loading reports missing keys, we map the backbone's \textsf{model.encoder} and \textsf{model.embeddings} prefixes to \textsf{encoder} and \textsf{embeddings}; we also map \textsf{hidden\_states\_norms.$i$} to \textsf{hidden\_states\_norms.stage}$j$, where $j=i+1$. We then load the corresponding missing tensors from the checkpoint, checking both key availability and tensor shapes.

\paragraph{Training inputs and losses.}
Training frames are fitted to the $2048{\times}1024$ reference canvas using a uniform scale, then rescaled by a random factor in $[0.5,2.0]$. We randomly crop $1024{\times}1024$ for SegFormer and $512{\times}1024$ for Mask2Former, padding as needed. Images undergo random horizontal flipping and photometric distortion before normalization with RGB mean $(0.485,0.456,0.406)$ and standard deviation $(0.229,0.224,0.225)$. Each of brightness, contrast, saturation and hue is perturbed independently with probability 0.5; brightness uses $[-32,32]$, contrast and saturation $[0.5,1.5]$, and hue integer shifts $[-18,18]$ in OpenCV's 8-bit HSV representation. Labels use the 19-class train-ID encoding, with 255 ignored. The models use their built-in training losses. For Mask2Former, the target supplies one binary mask for each class present in the crop; ignored pixels belong to none.

\paragraph{Optimization.}
All updates use AdamW with $(\beta_1,\beta_2)=(0.9,0.999)$ and mixed-precision training. We set the PyTorch seed to 0 and sample training images with replacement. For SegFormer, the base learning rate is $6{\times}10^{-5}$ and parameters whose name contains \emph{head} use ten times that rate. The weight decay is 0.01, except parameters whose name contains \emph{norm} and not \emph{head}, which have zero decay. The learning-rate multiplier is $(1-t/T)\min\{1,10^{-6}+(1-10^{-6})t/1500\}$ at iteration $t$, with $T=10{,}000$. For Mask2Former, the base rate is $10^{-4}$ and the \textsf{pixel\_level\_module.encoder} parameters use one tenth of that rate. Weight decay is 0.05 except for normalization and embedding parameters, which have zero decay. The learning-rate multiplier is $(1-t/T)^{0.9}$, with no warm-up. Mask2Former gradients are clipped to norm 0.01.

\paragraph{Interpolated candidate.}
Let $\theta_0$ and $\theta_{\rm FT}$ be the original and fine-tuned state dictionaries. For floating-point entries we form $\theta_\gamma=(1-\gamma)\theta_0+\gamma\theta_{\rm FT}$. At $\gamma=0$ and $1$ the original and fine-tuned states are used, respectively; for intermediate coefficients, non-floating-point entries are taken from the fine-tuned state. On training data we scan coefficients from 0 to 1 in steps of 0.1. Among candidates meeting the mean normal loss budget at $\tau=0.005$, we select the one with the highest adverse mIoU, breaking ties toward the smaller coefficient. Coefficient selection and scorer fitting are completed before the additional normal calibration images are used.

\clearpage\section{Learned whole-image selection on BDD and external validation on RaidaR}\label{sec:raidar}

\paragraph{Frames and calibration.}
We used all 3,779 sunny and 3,642 rainy image--mask pairs in the official segmentation test split, with the released semi-automatic masks. We mapped labels to the 19 Cityscapes classes, ignored other labels and resized images and masks to $720\times1280$ using bilinear and nearest-neighbor interpolation, respectively. We grouped consecutive sunny frame IDs across training and test splits. After excluding training frames in the two blocks containing both splits, we took the middle labeled frame of each remaining training block (21 images). All rules used the same 50 draws of ten images without replacement (seed 33427).

\paragraph{Frozen updates and selection rules.}
We kept the original WI coefficients (B2: 0.9, B0: 0.8, M2F: 0.7). Whole, the $8\times8$ grid and RALC use the original image scorer. The two local rules use the per-image bound. They rank cells by the mean confidence gap of their disagreeing pixels, or rank those pixels individually. Learned Whole changes only Whole's image ranking. We fitted it on the original 699 development images using ten features derived from image statistics, confidence, entropy and the disagreement fraction. All scorers stayed fixed on both test sets.

\paragraph{Mask2Former confidence.}
For each frozen model, we softmaxed each query's class logits including the no-object class, then discarded that class. For semantic class $c$ we summed $v_c=\sum_q p_q(c)\,\mathrm{sigmoid}(m_q)$ over queries, where $m_q$ is a mask logit. We bilinearly resized $v$ to one-quarter of the padded input resolution if needed, and formed log scores $\ell_c=\log[\max(v_c,10^{-8})/\max(\sum_j v_j,10^{-8})]$. We bilinearly upsampled the log scores to the padded input size in FP32, cropped to the image size and applied softmax over the 19 classes. The largest probability gives the confidence for that model's predicted label in Eq.~(2). Both resizes used \textsf{align\_corners=False}. The image scorer uses mean entropy and mean confidence at the lower resolution.

\paragraph{Training the whole-image selector.}
For each class $c$, let $T_c$ and $U_c$ be the original true positives and union pooled over adverse images in the training fold, and let $\mathcal C=\{c:U_c>0\}$. Updating all predictions on adverse image $i$ changes those counts by $\Delta T_{ic}$ and $\Delta U_{ic}$. Its training target is the first-order change in pooled mIoU (on the 0--1 scale):
\[
 b_i=\frac{1}{|\mathcal C|}\sum_{c\in\mathcal C}
 \left(\frac{\Delta T_{ic}}{U_c}-\frac{T_c\,\Delta U_{ic}}{U_c^2}\right).
\]
On normal images the harm target is the positive net error increase under full adoption divided by image area. We set benefit to zero on normal images and harm to zero on adverse images. We divided each target by its training standard deviation, centered it for the fit and restored its mean at prediction. Validation and testing used mIoU computed from pooled counts.

We standardized each input feature on the training fold. The RBF kernel was $K(z,z')=\exp[-\|z-z'\|^2/(2\sigma^2)]$, with $\sigma=k\sqrt{m}$, where $m$ is the median training-pair squared distance above $10^{-8}$. Ridge regression used $K+\lambda I$; images were ranked by predicted benefit minus $w$ times predicted harm. Five-fold validation grouped by scene selected $(k,\lambda,w)$ by adverse gain subject to the normal budget. The selected triples for B2, B0 and M2F were $(2,1,1)$, $(1,0.1,1)$ and $(0.5,0.01,0.5)$, respectively. Refitting on all 699 images gave $\sigma=7.125$, $3.592$ and $1.737$ in standardized feature space.

\clearpage
\paragraph{Learned whole-image selection on BDD at matched loss.}
On the original BDD pool we interpolated each rule's curve post hoc at a mean normal loss of 1,000 pixels/image. RALC minus Learned Whole was 2.428 pp on B2, 1.336 on B0 and 0.945 on M2F. The respective single-calibration 95\% paired intervals were $[1.692,3.599]$, $[0.756,2.142]$ and $[0.104,2.025]$. Each of 1,000 paired replicates used one common calibration draw, resampled normal and adverse test images within their pools, and re-matched both rules.

\paragraph{Fixed budget and budgets transferred from BDD.}
The primary comparison uses $\tau=0.005$ for every rule ($\alpha\approx9{,}694$ pixels). For transfer, we took the lower of the two nominal BDD budgets bracketing 1,000 loss pixels/image for each model and rule, then recalibrated at that budget on RaidaR. Table~\ref{tab:raidar-points} gives the resulting sunny losses. For each draw we pooled rainy confusion counts to compute mIoU gain over the original and averaged the per-image positive net error increases on sunny frames. We report means over the 50 draws.

\paragraph{Update endpoints.}
Original and fully updated rainy mIoU were 32.362 and 35.685 on B0, 37.836 and 39.264 on B2, and 39.486 and 39.450 on M2F. At the fixed budget, B2's learned whole-image rule captured almost all of RALC's gain. On M2F, the full update slightly degraded the original, leaving little absolute gain under either local rule.

\begin{table}[!htbp]\centering\small\caption{Rainy mIoU gain over the original model (pp) and mean sunny loss (pixels per image) on RaidaR.}\label{tab:raidar-points}
\begin{tabular}{llrr rrr}\toprule
Model & Rule & \multicolumn{2}{c}{Common budget $\tau=0.005$} & \multicolumn{3}{c}{Budget transferred from BDD} \\
\cmidrule(lr){3-4}\cmidrule(lr){5-7}
 & & Rainy gain & Sunny loss & $\tau$ & Rainy gain & Sunny loss \\ \midrule
B0 & Whole & 1.247 & 3,860 & 0.00475 & 1.135 & 3,694 \\
 & Learned Whole & 0.182 & 1,938 & 0.005 & 0.182 & 1,938 \\
 & $8\times8$ grid & 2.832 & 4,082 & 0.00225 & 2.247 & 1,507 \\
 & RALC & 2.881 & 4,093 & 0.0025 & 2.514 & 1,791 \\
\addlinespace
B2 & Whole & 1.039 & 3,220 & 0.005 & 1.039 & 3,220 \\
 & Learned Whole & 1.673 & 3,253 & 0.00475 & 1.664 & 3,204 \\
 & $8\times8$ grid & 1.693 & 3,188 & 0.00375 & 1.930 & 2,633 \\
 & RALC & 1.711 & 3,178 & 0.00375 & 1.988 & 2,603 \\
\addlinespace
M2F & Whole & -0.057 & 2,445 & 0.00575 & -0.038 & 2,526 \\
 & Learned Whole & -0.067 & 2,019 & 0.00575 & -0.060 & 2,265 \\
 & $8\times8$ grid & 0.036 & 2,490 & 0.003 & 0.292 & 1,552 \\
 & RALC & 0.041 & 2,489 & 0.00325 & 0.310 & 1,756 \\
\bottomrule\end{tabular}\end{table}

\clearpage
\paragraph{Uncertainty across test frames and calibration draws.}
For each of 1,000 paired bootstrap replicates (seed 33427), we resampled the five sunny and two rainy test blocks within condition and used the same blocks for all rules. We recomputed outcomes for all 50 calibration draws and averaged them for the draw-average interval; the single-draw interval used one common draw per replicate. Tables~\ref{tab:raidar-gain} and~\ref{tab:raidar-loss} give the 2.5th and 97.5th percentiles. The blocks follow frame numbers. Trip identifiers were unavailable, and the two rainy blocks provide little information about variation across routes.

On B2, RALC exceeded the learned whole-image rule in 25 of the 50 fixed-budget draws. Its single-draw gain interval also included zero on M2F. When budgets selected on BDD were applied to RaidaR, RALC had higher mean rainy gain and lower mean sunny loss than learned whole-image adoption on all three models. Against the grid, RALC spent more on B0 and M2F at the transferred budgets (Table~\ref{tab:raidar-points}).

\begin{table}[!htbp]\centering\small\caption{Paired differences in rainy mIoU gain (pp).}\label{tab:raidar-gain}
\begin{tabular}{llrrrr}\toprule
Model & Comparator & Budget & RALC minus comparator & Draw-average interval & Single-draw interval \\ \midrule
B0 & Learned Whole & Fixed & 2.699 & $[1.903,\, 3.098]$ & $[1.594,\, 3.427]$ \\
 &  & Transfer & 2.332 & $[1.490,\, 2.963]$ & $[1.252,\, 3.211]$ \\
 & $8\times8$ grid & Fixed & 0.050 & $[0.031,\, 0.050]$ & $[0.025,\, 0.059]$ \\
 &  & Transfer & 0.267 & $[0.203,\, 0.340]$ & $[0.189,\, 0.352]$ \\
\addlinespace
B2 & Learned Whole & Fixed & 0.038 & $[0.038,\, 0.205]$ & $[-0.140,\, 0.345]$ \\
 &  & Transfer & 0.324 & $[0.324,\, 0.373]$ & $[0.035,\, 0.630]$ \\
 & $8\times8$ grid & Fixed & 0.018 & $[0.018,\, 0.020]$ & $[-0.013,\, 0.068]$ \\
 &  & Transfer & 0.058 & $[0.051,\, 0.067]$ & $[0.015,\, 0.105]$ \\
\addlinespace
M2F & Learned Whole & Fixed & 0.109 & $[0.043,\, 0.225]$ & $[-0.029,\, 0.309]$ \\
 &  & Transfer & 0.370 & $[0.236,\, 0.477]$ & $[0.065,\, 0.689]$ \\
 & $8\times8$ grid & Fixed & 0.005 & $[0.004,\, 0.006]$ & $[-0.003,\, 0.044]$ \\
 &  & Transfer & 0.018 & $[-0.002,\, 0.048]$ & $[-0.111,\, 0.152]$ \\
\bottomrule\end{tabular}\end{table}

\begin{table}[!htbp]\centering\small\caption{Paired differences in mean sunny loss (pixels/image). Positive values indicate greater regression under RALC.}\label{tab:raidar-loss}
\begin{tabular}{llrrrr}\toprule
Model & Comparator & Budget & RALC minus comparator & Draw-average interval & Single-draw interval \\ \midrule
B0 & Learned Whole & Fixed & 2,155.5 & $[968.3,\, 2,663.2]$ & $[137.9,\, 3,504.8]$ \\
 &  & Transfer & -146.4 & $[-1,368.2,\, 566.6]$ & $[-2,287.5,\, 1,406.5]$ \\
 & $8\times8$ grid & Fixed & 11.1 & $[5.0,\, 13.9]$ & $[-2.5,\, 33.0]$ \\
 &  & Transfer & 284.3 & $[223.5,\, 400.7]$ & $[144.8,\, 573.5]$ \\
\addlinespace
B2 & Learned Whole & Fixed & -75.2 & $[-97.6,\, -23.4]$ & $[-399.8,\, 119.2]$ \\
 &  & Transfer & -600.5 & $[-741.7,\, -378.5]$ & $[-1,394.1,\, -6.4]$ \\
 & $8\times8$ grid & Fixed & -10.4 & $[-18.3,\, 0.7]$ & $[-39.4,\, 7.4]$ \\
 &  & Transfer & -30.0 & $[-54.3,\, 7.4]$ & $[-97.8,\, 22.8]$ \\
\addlinespace
M2F & Learned Whole & Fixed & 470.5 & $[17.7,\, 657.7]$ & $[0.0,\, 1,011.8]$ \\
 &  & Transfer & -509.1 & $[-1,239.1,\, -199.0]$ & $[-2,146.1,\, -58.3]$ \\
 & $8\times8$ grid & Fixed & -0.2 & $[-1.7,\, 2.4]$ & $[-11.9,\, 18.6]$ \\
 &  & Transfer & 204.0 & $[110.8,\, 464.5]$ & $[50.4,\, 564.2]$ \\
\bottomrule\end{tabular}\end{table}

\clearpage
\paragraph{Post hoc matching on RaidaR.}
We used the RaidaR sunny test labels to align each rule at the same measured loss. We evaluated the 77 nominal budgets from Sec.~S5 and averaged per-image sunny loss and pooled rainy mIoU gain over the 50 calibration draws at each budget. For each rule we interpolated between the first pair of adjacent budgets bracketing the target loss. Table~\ref{tab:raidar-local-curves} gives the gains at 1,000 pixels/image and three neighboring targets. Both local rules use the per-image bound.

The learned selector's adoption rate is the fraction of rainy images on which it accepts the full update, averaged across draws. At $\tau=0.005$ the rates were 91.3\%, 29.7\% and 76.2\% for B2, B0 and M2F, respectively. At 1,000 sunny loss pixels/image the corresponding interpolated rates were 14.9\%, 6.8\% and 32.7\%, using the same weights as for the gain. On B2, the share of RALC's advantage recovered by the grid was $(1.781-0.070)/(1.912-0.070)\approx93\%$. Learned Whole is the reference in both differences.

\begin{table}[!htbp]\centering\small\caption{Rainy mIoU gain over the original model (pp) at four measured sunny losses on RaidaR. Targets are mean loss pixels per sunny image.}\label{tab:raidar-local-curves}
\begin{tabular}{lrrrr}\toprule
Model & Target & Learned Whole & $8\times8$ grid & RALC \\ \midrule
B2 & 500 & 0.023 & 1.274 & 1.529 \\
 & 1,000 & 0.070 & 1.781 & 1.912 \\
 & 1,500 & 0.169 & 1.968 & 2.078 \\
 & 2,000 & 0.395 & 2.021 & 2.105 \\
\addlinespace
B0 & 500 & 0.015 & 1.149 & 1.485 \\
 & 1,000 & 0.055 & 1.930 & 2.086 \\
 & 1,500 & 0.113 & 2.243 & 2.379 \\
 & 2,000 & 0.217 & 2.455 & 2.564 \\
\addlinespace
M2F & 500 & -0.005 & 0.142 & 0.176 \\
 & 1,000 & -0.020 & 0.238 & 0.275 \\
 & 1,500 & -0.051 & 0.291 & 0.322 \\
 & 2,000 & -0.067 & 0.240 & 0.265 \\
\bottomrule\end{tabular}\end{table}

For Table~\ref{tab:raidar-local-pairs}, each paired bootstrap replicate resampled sunny and rainy frame blocks separately and used one common calibration draw for all rules. We then re-matched each rule at 1,000 pixels/image on the resampled sunny data and compared rainy gains. All 1,000 replicates bracketed the target. The intervals reflect the two observed rainy blocks and variation across calibration draws.

\begin{table}[!htbp]\centering\small\caption{Paired differences in rainy mIoU gain (pp) at a matched mean sunny loss of 1,000 pixels/image on RaidaR. Intervals are the 2.5th and 97.5th percentiles for a single calibration draw.}\label{tab:raidar-local-pairs}
\begin{tabular}{lrrrrrr}\toprule
Model & \multicolumn{2}{c}{RALC minus Learned Whole} & \multicolumn{2}{c}{Grid minus Learned Whole} & \multicolumn{2}{c}{RALC minus grid} \\ 
 & Difference & 95\% interval & Difference & 95\% interval & Difference & 95\% interval \\ \midrule
B2 & 1.842 & $[0.990,\, 2.281]$ & 1.711 & $[0.884,\, 2.126]$ & 0.131 & $[0.068,\, 0.202]$ \\
B0 & 2.031 & $[1.143,\, 2.959]$ & 1.874 & $[0.942,\, 2.796]$ & 0.157 & $[0.088,\, 0.253]$ \\
M2F & 0.295 & $[0.026,\, 0.667]$ & 0.258 & $[-0.020,\, 0.635]$ & 0.037 & $[0.018,\, 0.063]$ \\
\bottomrule\end{tabular}\end{table}

\clearpage\section{Validation results and adoption-rule comparisons on BDD}

Development used 500 normal Cityscapes images and 406 adverse ACDC validation images. The MUSES public training and validation splits supplied 350 clear-day and 1,400 adverse RGB images for development evaluation only. Normal BDD images were dry daytime frames under clear, partly cloudy or overcast skies.

\paragraph{Reconstructing the development pool.}
The scorer pool is the same for all three architectures. For Cityscapes, sort the training filenames from Aachen and Bremen together and take 200 entries at indices $\lfloor iN/200\rfloor$, $i=0,\ldots,199$, from the list of length $N$. For ACDC, sort labeled training frames by sequence and filename within each condition. Form the clean-reference list by concatenating the fog, night, rain and snow lists in that order. Take 200 evenly spaced entries from each of the five lists before keeping only sequences in group B of Sec.~S11. In that sequence split, process descending sequence sizes with name ties in lexicographic order; assign a sequence to A when the two groups have equal sizes. This yields 200 clean Cityscapes images and 499 ACDC images: 100 clean references, 101 fog, 99 night, 100 rain and 99 snow. Paired ACDC reference and adverse images share a scene name, giving 699 images from 599 scene identities.

\paragraph{Reference pixel budget and freezing order.}
The 300 clean images in this development pool also define the reference count $P_{\rm ref}=1{,}938{,}835.2467$: their total of 581,650,574 valid pixels divided by 300. We count only the 19 train-ID classes and exclude label 255. Thus the reference set is independent of BDD calibration and testing, while sharing the scorer's normal training images. We first train the update on group A, then select its interpolation coefficient and fit the scorer using the development pool. These models, the reference budget and the candidate construction are fixed before drawing BDD calibration images. The original BDD pool and draws are reconstructed in Sec.~S9.

\paragraph{Inference and recorded environment.}
Segmentation uses native-resolution RGB images, without test-time augmentation or resizing. After the normalization in Sec.~S11, reflection padding on the right and bottom makes the dimensions multiples of 32. We bilinearly upsample class scores to the padded size with \textsf{align\_corners=False}, remove the padding and apply softmax to obtain the selected-label confidence; image entropy and mean confidence are computed at the score-map resolution before upsampling. Sec.~S12 describes Mask2Former's semantic score construction. The archived segmentation environment is Python 3.12.3, PyTorch 2.5.1 with CUDA 12.4, Transformers 4.46.3, NumPy 2.1.3 and Pillow 11.0.0. Frozen prediction records use FP32 without autocast or TF32 matrix multiplication.

\paragraph{Checkpoint identification.}
The release names in Sec.~S11 are identified by the following SHA-256 hashes of the initial and final fine-tuned weight files. Interpolated states are reconstructed using the coefficients in the main paper.
\begin{flushleft}\footnotesize
\textsf{B2 initial: 584d72ace8c17c5683c198e1a4c3eda2e1026bafc02711ac5e5a4dc097bd05fe}\\
\textsf{B2 FT: e74ca7706830e75b0e848253a1fb42cee397b6be9d12f8da4be2260ff7ba4a96}\\
\textsf{B0 initial: 027ed78d8ff9c535df5a66361c92b9673cca3cb923cbeb8c5802385b6b93194c}\\
\textsf{B0 FT: 4c6af9b4363703e0fdcd3fdeec03ba5d3ac951bb7842141ea8cc4ae2f4c96fb0}\\
\textsf{M2F initial: 8f0e42a9c3d4f5a2a628b042a065c94b6acfa4f202340c186a89c9741755a7ac}\\
\textsf{M2F FT: 74d279af7f48c5132ead9efe152694151cb9b2369dc34dbb0cca047ccf0edcd5}\\
\end{flushleft}

\paragraph{Weather subsets.}
At the fixed budget, RALC exceeded Whole and the $32{\times}32$ grid in rain, snow and dry night on all three WI updates. Only two BDD fog images qualified.

\paragraph{Overall mIoU on a separate validation split.}
On the official BDD100K validation split, we measured overall mIoU on 998 aligned image--mask pairs across scene conditions. None overlapped the original pool. Both grid and pixel rules used the original 50 ten-image calibration draws (Sec.~S1).

\begin{table}[!htbp]
\centering
\caption{Fixed budget ($n=10$, $\tau=0.005$). Overall mIoU (\%) uses the 998 official validation images. Risk ratios $\rho$ use the original 250 held-out normal images. Grid cells are $8{\times}8$ pixels. Pixel--grid differences use unrounded mIoU.}
\label{tab:official}
\begingroup\small\setlength{\tabcolsep}{3pt}\centering
\begin{tabular*}{\textwidth}{@{\extracolsep{\fill}}lrrrrrrrrrr@{}}
\toprule
 & \multicolumn{4}{c}{Shared bound} & \multicolumn{4}{c}{Per-image bound} & \multicolumn{2}{c}{Pixel minus grid (pp)} \\
\cmidrule(lr){2-5}\cmidrule(lr){6-9}\cmidrule(l){10-11}
 & \multicolumn{2}{c}{Grid} & \multicolumn{2}{c}{Pixel} & \multicolumn{2}{c}{Grid} & \multicolumn{2}{c}{Pixel} & Shared & Per-image \\
Model & mIoU & $\rho$ & mIoU & $\rho$ & mIoU & $\rho$ & mIoU & $\rho$ & & \\
\midrule
B2 & 53.423 & 0.020 & 53.543 & 0.020 & 55.944 & 0.147 & 55.977 & 0.149 & 0.120 & 0.033 \\
B0 & 46.500 & 0.107 & 46.574 & 0.108 & 48.083 & 0.281 & 48.115 & 0.282 & 0.074 & 0.033 \\
M2F & 57.397 & 0.038 & 57.452 & 0.037 & 57.943 & 0.161 & 57.978 & 0.161 & 0.054 & 0.035 \\
\bottomrule
\end{tabular*}\endgroup

\end{table}
With the per-image bound, refining cells to pixels added 0.033--0.035 pp (Table~\ref{tab:official}). All three pixel--grid intervals excluded zero for the mean over calibration draws and included zero for single deployments (Table~S2).

\paragraph{Comparison rules.}
Whole accepts or rejects all disagreements together. It uses the same scorer as both pixel variants, sums full-update losses on calibration images scoring at least $s(x)$, adds $|D(x)|$, and accepts if the total is at most $(n+1)\alpha$. A learned alternative, ReCIRC-instance, uses the same per-image count as default RALC. It inverts each image's predicted loss curve at common predicted-risk levels to construct its candidates. We train the gradient-boosted regressor on 300 normal images from the scorer training pool before BDD calibration (Sec.~S8). ReCIRC-shared and ReCIRC-global use the same learned family with a shared or uniform reserve. At the primary budget with ten calibration images, the global reserve of 921,600 pixels forces rollback.

Region-Conf.\ ranks 8-connected disagreement components by mean confidence gap. The grid rules group disagreements in $8{\times}8$ or $32{\times}32$ cells aligned to the image origin. Region-Entropy adapts the entropy-based regional ordering of Wang et al. (main paper, Ref.~[\WangRef]) using normalized entropy differences. This ordering control uses our net-error loss and count-bound calibration. Pixel-Entropy ranks pixels by the normalized entropy decrease. Random pixel order assigns independent uniform scores before binning. We average its results over three seeds per draw. These controls share the image priority and calibration protocol, and each uses nested candidate sets. Each has its own calibration loss envelope.

\begin{table}[!htbp]
\caption{Task-adapted learned risk curves on BDD, WI updates, $n=10$ and $\tau=0.005$. Entries are adverse gain (pp) and mean normal risk ratio.}
\label{tab:fullrecirc}
\centering
{\centering\small\setlength{\tabcolsep}{1.8pt}
\begin{tabular*}{\columnwidth}{@{\extracolsep{\fill}}lrrrrrr@{}}
\toprule
Method & \multicolumn{2}{c}{B2} & \multicolumn{2}{c}{B0} & \multicolumn{2}{c}{M2F} \\
 & $\Delta_{\rm adv}$ & $\rho$ & $\Delta_{\rm adv}$ & $\rho$ & $\Delta_{\rm adv}$ & $\rho$ \\
\midrule
RALC & 4.427 & 0.149 & 3.166 & 0.282 & 2.114 & 0.161 \\
ReCIRC-global & 0.000 & 0.000 & 0.000 & 0.000 & 0.000 & 0.000 \\
ReCIRC-shared & 0.215 & 0.019 & 0.030 & 0.006 & 0.200 & 0.015 \\
ReCIRC-instance & 1.341 & 0.113 & 0.525 & 0.122 & 0.709 & 0.113 \\
\bottomrule
\end{tabular*}
}
\end{table}

\begin{table}[!htbp]
\caption{Adoption granularity and ordering on the original BDD test pool. Paired columns give adverse gain (pp) and normal risk ratio for the WI updates at the fixed budget in the main paper (Table~1).}
\label{tab:fullordering}
\centering
{\centering\small\setlength{\tabcolsep}{1.8pt}
\begin{tabular*}{\columnwidth}{@{\extracolsep{\fill}}lrrrrrr@{}}
\toprule
Controller & \multicolumn{2}{c}{B2} & \multicolumn{2}{c}{B0} & \multicolumn{2}{c}{M2F} \\
 & $\Delta_{\rm adv}$ & $\rho$ & $\Delta_{\rm adv}$ & $\rho$ & $\Delta_{\rm adv}$ & $\rho$ \\
\midrule
Whole & 1.235 & 0.103 & 0.492 & 0.110 & 0.225 & 0.068 \\
Region-Conf. & 2.533 & 0.123 & 1.690 & 0.220 & 1.364 & 0.139 \\
$32{\times}32$ grid & 4.182 & 0.145 & 3.054 & 0.291 & 2.012 & 0.161 \\
RALC & \textbf{4.427} & 0.149 & \textbf{3.166} & 0.282 & \textbf{2.114} & 0.161 \\
\midrule
Region-Entropy & 2.418 & 0.128 & 1.565 & 0.203 & 1.333 & 0.131 \\
Pixel-Entropy & 4.244 & 0.137 & 2.934 & 0.275 & 1.965 & 0.165 \\
Random pixel order & 3.631 & 0.212 & 2.249 & 0.367 & 1.295 & 0.186 \\
Shared (pixels) & 2.771 & 0.020 & 2.281 & 0.108 & 1.015 & 0.037 \\
\bottomrule\end{tabular*}
}
\end{table}

\paragraph{Ordering the changes.}
On the original BDD pool at the fixed budget, confidence ordering retained 0.796, 0.917 and 0.819 pp more adverse gain than random pixel ordering on B2, B0 and M2F, with lower mean normal loss on all three (Table~\ref{tab:fullordering}). Random scores change both the order and the number of pixels in each bin. Pixel-Entropy was closer, trailing by 0.183, 0.233 and 0.149 pp.

\paragraph{Image priority.}
We removed the image scorer by applying the same confidence-gap cutoffs to every image. We recalibrated this family on the same draws and kept the per-image bound. At $\tau=0.005$ the simpler rule gained 0.015, 0.051 and 0.145 pp over RALC on the adverse test images (B2, B0 and M2F). At matched loss, all three paired intervals included zero on the adverse images. At the same matched loss, removing the scorer raised B0 overall validation mIoU by 0.312 pp (95\% interval 0.159--0.426; Table~S8).

\paragraph{Per-image and shared bounds.}
At the fixed budget, the shared bound gained 0.790--1.790 pp over Whole on WI. Using the per-image bound instead added 0.885--1.656 pp and increased measured normal loss on all three models.

At matched loss the shared pixel rule gained 1.070--3.578 pp over Whole. It exceeded default RALC by 1.150 pp on B2 (paired 95\% interval 0.440--2.002) and 0.362 pp on B0, but trailed by 0.065 pp on M2F. The B0 and M2F intervals included zero (Table~S6).

On the official validation split, the per-image bound on $8{\times}8$ cells raised overall mIoU by 0.545--2.521 pp at the fixed budget, with higher normal risk (Table~\ref{tab:official}). At matched risk, the shared versus per-image intervals included zero for grids and pixels on all three models (Table~S7).

\paragraph{Additional BDD results at larger calibration sizes.}
At $n=50$ and $\tau=0.005$, Whole and RALC retained 7.021 and 8.517 pp on B2. At the second prespecified operating point ($n=250$, $\tau=0.0025$), RALC and Whole retain the same WI gain on B2. RALC retains more on B0 and slightly less on M2F. Full B0 WI exceeds the budget ($\rho=1.312$). RALC attains $\rho=0.942$ and retains 5.253 pp. On B0 FT, RALC retains 4.586 pp while lowering the risk ratio from 2.222 under full adoption to 0.936.

With 250 calibration images, ReCIRC-instance ties RALC on all three WI updates at $\tau=0.005$ and gains 0.014 pp on M2F at $\tau=0.0025$.

Across 72 prespecified settings, the highest mean RALC risk ratio was 0.988. Reallocating the normal pool for B0 WI at $n=250$ and $\tau=0.0025$ gave a mean ratio of 0.849, with 24.0\% of reallocations above budget. Supplementary Sec.~S9 gives the FT results and the recording-level analysis on MUSES.

\clearpage
\paragraph{What do ten calibration images add to local adoption?}
We held the update, scorer, confidence bins and candidate sets fixed. With no additional labeled BDD images ($n=0$), the loss sum is zero and the largest allowed set has $|A(x,\lambda)|\leq\alpha$. This caps the number of replacements on every image, including positions without valid labels. With $n=10$, the same rule allows $|A(x,\lambda)|\leq11\alpha-S_{10}(\lambda)$ and controls expected normal loss. Both use $\tau=0.005$ on the original BDD split. Table~\ref{tab:zerocal} reports the resulting change in gain and measured loss.

\begin{table}[htbp]\centering\small
\caption{Effect of ten additional normal calibration images on default pixel adoption. Gain is measured on the 500 adverse images relative to the original model. Normal loss is mean positive net error increase on the 250 held-out normal images (pixels/image). $\Delta$ is the gain at $n=10$ minus the gain at $n=0$.}\label{tab:zerocal}
\begin{tabular}{lrrrrrcc}
\toprule
& \multicolumn{3}{c}{Adverse gain (pp)} & \multicolumn{2}{c}{Normal loss} & \multicolumn{2}{c}{Paired 95\% interval for $\Delta$ (pp)} \\
\cmidrule(lr){2-4}\cmidrule(lr){5-6}\cmidrule(lr){7-8}
Update & $n=0$ & $n=10$ & $\Delta$ & $n=0$ & $n=10$ & Draw mean & Single draw \\
\midrule
B2 WI & 0.603 & 4.427 & 3.824 & 42 & 1,446 & [3.272, 4.710] & [2.718, 4.831] \\
B0 WI & 0.431 & 3.166 & 2.735 & 151 & 2,730 & [2.393, 3.297] & [2.007, 3.409] \\
M2F WI & 0.489 & 2.114 & 1.625 & 126 & 1,560 & [0.765, 2.735] & [0.557, 2.911] \\
B0 FT & 0.397 & 2.783 & 2.386 & 144 & 3,001 & [2.066, 2.905] & [1.796, 3.010] \\
\bottomrule
\end{tabular}
\end{table}
The $n=0$ rule gives a single prediction per image. For $n=10$, we average gains over the original 50 calibration draws. For both intervals we resampled the 500 adverse images with replacement 1,000 times, using the same image sample for both rules and recomputing pooled mIoU. Each replicate either averages the gain difference over all 50 draws or uses one uniformly sampled draw.

\clearpage\section{What changes together inside an $8{\times}8$ cell?}
\label{sec:within-cell}
We examined the frozen WI updates on the 699 images used to develop the image scorer: 399 adverse ACDC images and 300 normal Cityscapes or ACDC images. We call a cell \emph{mixed} if at least one replacement corrects a wrong label and another overwrites a correct label. A change between two wrong labels satisfies neither condition. We pooled labeled disagreements within each scene condition and counted the fraction located in mixed cells.

On adverse images, mixed cells contained 3.31--3.90\% of labeled disagreements (Table~\ref{tab:cellmix}); on normal images the fraction was 12.80--15.68\%. Even inside mixed cells, the default confidence-gap ranking usually put a correction ahead of a newly introduced error. About four in five such pairs on adverse images had the same original and proposed classes. We therefore tested a class-transition adjustment within each confidence bin.

\begin{table}[ht]\centering
\caption{Within-cell structure on the development pool. Mixed share is the percentage of labeled disagreements in mixed cells. Ordering accuracy compares every correction with every newly introduced error in the same cell, using RALC's 32 confidence-gap bins and counting ties as half correct. The last column gives the percentage of adverse-scene pairs with the same old-to-new class transition.}
\label{tab:cellmix}
\begin{tabular}{lrrrrr}
\toprule
& \multicolumn{2}{c}{Mixed share (\%)} & \multicolumn{2}{c}{Ordering accuracy (\%)} & Same transition (\%) \\
\cmidrule(lr){2-3}\cmidrule(lr){4-5}
Model & Adverse & Normal & Adverse & Normal & Adverse \\
\midrule
SegFormer-B2 & 3.83 & 14.55 & 81.60 & 80.96 & 82.07 \\
SegFormer-B0 & 3.31 & 12.80 & 79.19 & 79.41 & 80.13 \\
Mask2Former  & 3.90 & 15.68 & 81.69 & 79.78 & 79.64 \\
\bottomrule
\end{tabular}
\end{table}

\paragraph{Does the class transition improve the ranking?}
Within each gap bin, we estimated the number of corrections minus newly introduced errors per labeled disagreement for each old-to-new class pair. We subtracted the bin-wide rate and shrank the difference toward zero by multiplying it by $N/(N+4096)$, where $N$ counts labeled disagreements for that pair and bin. We added the difference to the confidence gap, clipped the result to $[-1,1]$ and used the original 32 bins. For the grid rule, we averaged the scores of all disagreements in each cell before binning.

We split 217 source-and-recording groups into five folds and fitted both the class adjustment and the image scorer on the other four folds. In each held-out fold, all four rules used the same 50 draws of ten normal calibration images. Test images sharing a recording with a calibration image were excluded for that draw. Each rule used its own calibration losses and the per-image replacement bound. For each draw index, we pooled adverse confusion counts across held-out folds before computing mIoU.

At a post hoc matched mean normal loss of 1,000 pixels per image, the adjustment lowered adverse mIoU gain for both pixels and cells on all three models (Table~\ref{tab:pairrank}). At the fixed budget ($\tau=0.005$), the adjusted pixel rule also retained less gain and incurred higher measured normal loss on every model. RALC uses the original confidence-gap ranking.

\begin{table}[ht]\centering
\caption{Adverse mIoU gain over the original model (pp), matched to the same measured normal loss on the development pool. $\Delta$ is adjusted minus original pixel gain. The bracketed values are the 2.5th and 97.5th percentiles across 50 paired calibration repeats, with loss re-matched within each repeat.}
\label{tab:pairrank}
\begin{tabular}{lrrrrrc}
\toprule
& \multicolumn{2}{c}{Original gap} & \multicolumn{2}{c}{Class-adjusted gap} & \multicolumn{2}{c}{Pixel difference} \\
\cmidrule(lr){2-3}\cmidrule(lr){4-5}\cmidrule(lr){6-7}
Model & Pixel & Grid & Pixel & Grid & $\Delta$ & Paired percentiles \\
\midrule
SegFormer-B2 & 2.871 & 2.538 & 2.126 & 1.840 & $-0.745$ & $[-1.264,-0.240]$ \\
SegFormer-B0 & 2.031 & 1.779 & 1.476 & 1.270 & $-0.555$ & $[-0.803,-0.246]$ \\
Mask2Former  & 4.320 & 4.055 & 3.496 & 3.262 & $-0.824$ & $[-1.236,-0.435]$ \\
\bottomrule
\end{tabular}
\end{table}

\clearpage\section{Candidate construction and additional bound properties}
\label{sec:method-details}
\paragraph{Confidence bins and shared levels.}
Let $r_p$ be the confidence gap and $s(x)$ the fixed image score defined in the main paper. The following definitions instantiate the nested candidates used in the experiments.
Partition $[-1,1]$ into $K=32$ equal bins and assign pixel $p$ to $k_p=\min(K-1,\lfloor K(r_p+1)/2\rfloor)$. At each shared level $\lambda\in\{-K,\ldots,3K\}$, a higher image score can lower the bin threshold. With $c(u)=\min(K,\max(0,u))$ and $q(x)=\operatorname{round}(2K\min(1,\max(0,2s(x)-1)))$, the threshold is
\begin{equation}
 b(x,\lambda)=
 \begin{cases}
 c(\lambda-q(x)), & s(x)\geq 0.5,\\
 c(K+\lambda), & s(x)<0.5 .
 \end{cases}
\tag{S15.1}\label{eq:bin-threshold}
\end{equation}
The candidate at level $\lambda$ contains the disagreeing pixels with $k_p\geq b(x,\lambda)$. Raising $\lambda$ can only remove pixels. The first level contains all disagreements and the last contains none. The 129 shared levels let us sum calibration losses at the same level across images. On one image they yield at most 33 distinct masks, whose sizes follow from the bin counts.

\paragraph{Computing the calibration envelope.}
For each calibration image, scan the candidate levels from the empty set toward full adoption and retain the largest loss seen so far. Summing these running maxima across images gives the stored calibration loss sum at each level.

\paragraph{Tightness and unused budget.}
The count bound is tight when complete ground-truth labels are unrestricted. If $y=f_0(x)$, every adopted change adds an error. Discrete bins can leave some budget unused. When one level passes and its predecessor fails, the unused allowance is smaller than the drop in $S_n+B$ between them.

\clearpage\section{Complete qualitative cases and selection records}
\label{sec:qualitative-cases}
\begin{figure}[htbp]
\centering\includegraphics[width=\linewidth]{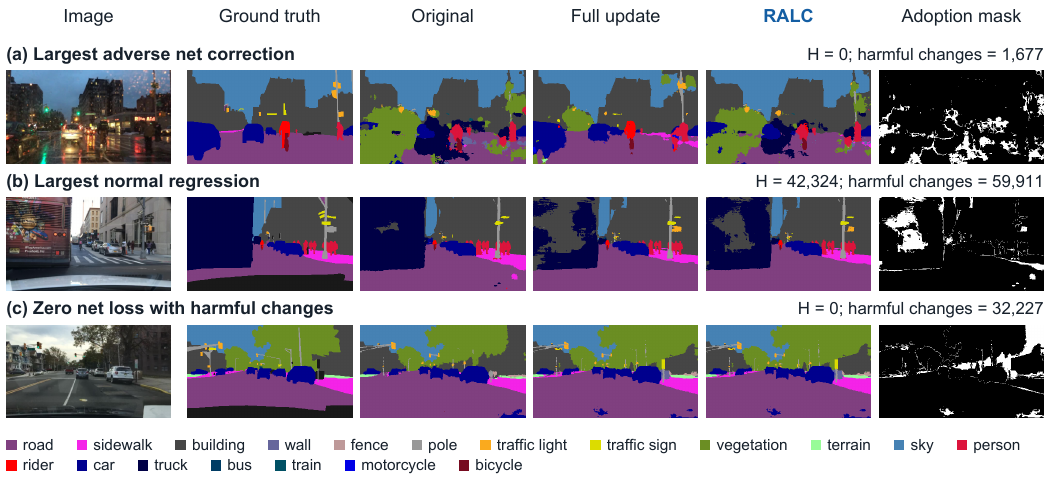}
\caption{Complete archived predictions for the three B2 WI cases. White marks adopted changes. Case (a) also appears as a registered crop in the main paper. Case (c) has zero net regression loss despite 32,227 newly introduced errors, offset by 36,285 corrections. Counts refer to complete images.}
\label{fig:full-cases}
\end{figure}

These are the archived B2 WI cases from the first calibration draw, with $n=10$ and $\tau=0.005$. The selection contains the adverse image with the largest net correction, the normal image with the largest regression loss $H$, and, among normal images with $H=0$, the image with the most newly introduced errors. Figure~\ref{fig:full-cases} retains the complete fields of view. The main paper enlarges two locations in case (a) to show an adopted correction and a correct original label retained despite a harmful proposal. Cases (b) and (c) document the normal-scene errors reported in the main text.

\begin{table}[htbp]\centering\small
\caption{Existing full-resolution replay records for the cases above. The displayed maps are downsampled; all counts below come from the original records. $H$ is the positive part of newly introduced errors minus corrections.}
\label{tab:case-records}
\begin{tabular}{llrrrr}
\toprule
Case & BDD image ID & Corrections & New errors & $H$ & Adopted changes \\
\midrule
(a) & \mbox{196d16dd-e26c2301} & 101,176 & 1,677 & 0 & 106,282 \\
(b) & \mbox{a822c459-20b67e7f} & 17,587 & 59,911 & 42,324 & 83,291 \\
(c) & \mbox{ab0735fb-f491ca69} & 36,285 & 32,227 & 0 & 84,860 \\
\bottomrule
\end{tabular}
\end{table}

\clearpage\section{Application to COCO multi-label recognition}\label{sec:coco_app}
We apply replacement-count control to 80 binary category-presence decisions per image, using category labels as adoption units.

\paragraph{Models and updates.}
ResNet-50 and ConvNeXt-Tiny start from torchvision ImageNet-1K weights and train on all 82,783 COCO2014 training images for 20 epochs (seed 16112026). A category is positive if any instance is annotated. Inputs use $224\times224$ resizing, random horizontal flips and ImageNet normalization. ASL has focusing exponents 4 for negatives and 0 for positives, clipping 0.05 and detached focusing weights. AdamW uses learning rates $10^{-4}$ for the backbone and $10^{-3}$ for the head, weight decay $10^{-4}$, one warm-up epoch and cosine decay to 0.05 of the initial rate. Head biases use training-category log odds with frequencies clipped to $[0.001,0.999]$.

Each classifier is then fine-tuned on all training images for 3 epochs with seeds 16112027 and 16112028, giving four fixed updates. Fine-tuning uses learning rate $2\times10^{-5}$, the same decay and cosine floor, and frozen batch-normalization statistics. Before normalization, apply $x\mapsto x^\gamma$, $\gamma\sim U[2,4]$, followed by $9\times9$ Gaussian blur with standard deviation $U[0.8,2]$. Test values are $\gamma=3$ and standard deviation 1.4. Training uses BF16 autocast, FP32 weights and batch size 64; inference is FP32. We use the final epoch.

\paragraph{Data splits and fixed decisions.}
We partition 40,504 validation images by original ID (seed 20261116) into 4,000 development, 2,000 calibration-pool, 10,000 clean-test and 24,504 corrupted-test images. All views of an image stay together. After the initial evaluation at a threshold of 0.5, we reanalyzed the frozen predictions. For each checkpoint, we selected a single threshold on the development set by maximizing pooled micro-F1 over equal numbers of clean and corrupted views. We searched $0.01,0.02,\ldots,0.99$; ties favor the value closest to 0.5, then the smaller value. Thresholds for the original and its two updates are $(0.76,0.84,0.85)$ for ResNet-50 and $(0.79,0.87,0.85)$ for ConvNeXt-Tiny, fixed before risk calibration.

For sigmoid score $p$ and development threshold $t$, center the score as
\[
q=\frac{p(1-t)}{p(1-t)+(1-p)t}.
\]
We checked that $q\geq0.5$ preserves every selected binary decision. On disagreements, rank the difference in the models' confidence in their own labels using 32 fixed bins on $[-1,1]$. Candidate sets take suffixes of these bins, including full rollback. The optional image scorer is constant.

\paragraph{Risk and evaluation.}
With $e(\hat y,y)=\sum_{c=1}^{80}\mathbf{1}\{\hat y_c\ne y_c\}$, use
\[
H(x)=\bigl[e(\hat y_A(x),y)-e(\hat y_0(x),y)\bigr]_+,
\qquad B(x,\lambda)=|A(x,\lambda)|.
\]
The same count bound and monotone-envelope calibration apply. Set $\alpha=0.25$ label errors per clean image. Take 10 images from each of 50 calibration orders, sampled without replacement within each order from the clean pool (seed 20261117). Utility is pooled micro-F1 on the corrupted test set; clean loss takes the positive net error increase per image before averaging.

Table~\ref{tab:coco_app} averages the 50 orders. Its intervals use 1,000 paired image-resampling replicates (seed 20261118), each selecting one calibration order. They include image and single-calibration-set variability. Individual calibration sets can have mean loss above budget. On corrupted images, the original ResNet-50 has micro-F1 57.622 and macro-F1 50.354; ConvNeXt-Tiny has 63.719 and 57.891, respectively.

\begin{table}[htbp]
\centering
\caption{COCO application with $n=10$ and $\alpha=0.25$. Gains are percentage points over the original classifier; mean loss is label errors per clean image. Brackets give 95\% image-and-calibration resampling intervals.}\label{tab:coco_app}
\small
\begin{tabular}{llrrr}
\toprule
Backbone & Update seed & Micro-F1 gain [interval] & Mean loss [interval] & Macro-F1 gain\\
\midrule
ResNet-50 & 16112027 & $7.133\,[0.995,11.135]$ & $0.147\,[0.006,0.271]$ & 8.213\\
ResNet-50 & 16112028 & $6.879\,[0.974,11.160]$ & $0.141\,[0.005,0.264]$ & 7.951\\
ConvNeXt-Tiny & 16112027 & $5.891\,[0.677,8.762]$ & $0.136\,[0.008,0.231]$ & 6.721\\
ConvNeXt-Tiny & 16112028 & $5.993\,[1.170,8.771]$ & $0.139\,[0.014,0.241]$ & 6.901\\
\bottomrule
\end{tabular}
\end{table}

The four updates retain positive micro-F1 and macro-F1 gains after risk-controlled label adoption. This application check extends the count-based procedure from pixel labels to category labels under the same expected-loss criterion.

\end{document}